\documentclass{fcs}
\usepackage{bm}
\usepackage{hyperref}       
\usepackage{times}  
\usepackage{helvet}  
\usepackage{courier}  
\usepackage{url}  
\usepackage{graphicx}  
\usepackage{algorithm,algorithmic}
\usepackage{amsmath,amssymb,amsthm}
\usepackage{graphicx}
\usepackage{subfigure} 
\usepackage{color}
\usepackage{booktabs}
\usepackage{footnote}
\usepackage{xspace}
\makesavenoteenv{tabular}
\makesavenoteenv{table}
\usepackage{pdflscape}
\usepackage{array}
\usepackage{booktabs}
\usepackage{diagbox}

\newcommand{\argmin}[1]{\mathop{\hbox{argmin}}_{#1}~\!}

\newcommand{\diag}[1]{\hbox{diag}(#1)}

\newcommand{\be}{\begin{equation}}
\newcommand{\ee}{\end{equation}}
\newcommand{\bea}{\begin{eqnarray}}
\newcommand{\eea}{\end{eqnarray}}
\newcommand{\ben}{\begin{equation*}}
\newcommand{\een}{\end{equation*}}
\newcommand{\bean}{\begin{eqnarray*}}
\newcommand{\eean}{\end{eqnarray*}}

\theoremstyle{plain}

\theoremstyle{definition}

\theoremstyle{remark}

\def\argmin{\mathop{\rm arg\,min}}

\def\norm#1{\left\|#1\right\|}

\def\mbra#1{\left[#1\right]}

\def\wada{\textsc{Wada}\xspace}
\def\WADA{\wada}
\def\wagmf{\textsc{Wagmf}\xspace}
\def\WAGMF{\wagmf}
\def\adam{\textsc{Adam}\xspace}
\def\Adam{\adam}
\def\rmsprop{\textsc{RmsProp}\xspace}
\def\RmsProp{\rmsprop}
\def\adagrad{\textsc{AdaGrad}\xspace}
\newcommand{\AdaGrad}{\adagrad}
\def\amsgrad{\textsc{AmsGrad}\xspace}
\def\AmsGrad{\amsgrad}
\def\adamnc{\textsc{AdamNc}\xspace}
\def\AdamNc{\adamnc}

\def\threeada{\textsc{Wada-v3}\xspace}
\def\fourada{\textsc{Wada-v4}\xspace}

\def\R{\mathbb{R}}
\def\E{\mathbb{E}}
\def\F{\mathcal{F}}

\def\citep{\cite} 
\def \emph#1{{\color{red}{#1}}} 

\volumn{ }
\doi{ }
\articletype{RESEARCH~ARTICLE}
\copynote{{\copyright} Higher Education Press and Springer-Verlag Berlin Heidelberg 2019}
\ratime{Received December 30, 2018; accepted May 10, 2019}
\email{cheneh@ustc.edu.cn}
\title{Adam revisited: a weighted past gradients perspective}
\author{Hui \MakeUppercase{Zhong} $^{1}$, Zaiyi \MakeUppercase{Chen} $^{2}$, Chuan \MakeUppercase{Qin} $^{1}$, Zai \MakeUppercase{Huang} $^{1}$, Vincent W. \MakeUppercase{Zheng}$^{3}$, Tong \MakeUppercase{Xu} $^{1}$,\\ Enhong \MakeUppercase{Chen} \xff $^{1}$}
\address{
  {1\quad Anhui Province Key Laboratory of Big Data Analysis and Application,\\ University of Science and Technology of China, Hefei 230027, China}\\
  {2\quad Zhejiang Cainiao Supply Chain Management Co. Ltd}\\
  {3\quad Advanced Digital Sciences Center, Singapore}
}

\begin{document}
  \maketitle
  \setcounter{page}{1}
  
  \setlength{\baselineskip}{14pt}
  
\begin{abstract}
Adaptive learning rate methods have been successfully applied in many fields, especially in training deep neural networks. 
Recent results have shown that adaptive methods with exponential increasing weights on squared past gradients (i.e., \textsc{Adam}, \textsc{RMSProp}) may fail to converge to the optimal solution.
Though many algorithms, such as \amsgrad and \adamnc, have been proposed to fix the non-convergence issues, achieving a data-dependent regret bound similar to or better than \AdaGrad is still a challenge to these methods.
In this paper, we 
propose a novel adaptive method \textit{Weighted Adaptive Algorithm} (\textsc{Wada}) to tackle the non-convergence issues.
Unlike \amsgrad and \adamnc, we consider using a milder growing weighting strategy on squared past gradient, in which weights grow linearly.
Based on this idea, we propose \textit{Weighted Adaptive Gradient Method Framework} (\textsc{Wagmf}) and implement \wada algorithm on this framework.
Moreover, we prove that \textsc{Wada} can achieve a weighted data-dependent regret bound,  which could be better than the original regret bound of \adagrad when the gradients decrease rapidly. This bound may partially explain the good performance of \adam in practice.
Finally, extensive experiments 
demonstrate the effectiveness of \WADA and its variants in comparison with several variants of \Adam on training convex problems and deep neural networks.
\end{abstract}


\section{Introduction}

Recently, many adaptive learning rate variants of stochastic gradient descent (SGD) algorithm~\cite{robbins1951stochastic} have been shown to be very successful in training neural networks, such as \AdaGrad~\citep{duchi2011adaptive}, \RmsProp~\citep{tieleman2012lecture}, \textsc{AdaDelta}~\citep{zeiler2012adadelta}, \Adam~\citep{kingma2014adam}. 
These methods can be applied to many training tasks, such as, text recognition~\citep{yin2018transcribing}, image ranking~\citep{krizhevsky2012imagenet}, online education~\citep{su2018exercise,liu2018finding}. Especially for tasks with sparse data, such as TF-IDF~\citep{salton1988term}, or deep neural networks, i.e., multi-layer perceptron, convolutional neural networks~\citep{lecun1998gradient}.
The idea of these adaptive methods is to adjust the learning rate for every parameter, by analyzing the gradients during past iterations.
\AdaGrad is the pioneer of these algorithms, and it achieves the well-known data-dependent regret bound $ O(\sum_{i=1}^{d} \norm{ g_{1:T,i} }_2 ) $, where $ T $ is the iteration number and $ g_{1:T,i} $ is a vector of historical  gradients of the $ i $-th dimension, in the case of training online convex problems.
The data-dependent regret bound can be better than the original  $ O(\sqrt{T}) $ regret bound, which is known to be optimal~\citep{Hazan:2016:IOC:3006427.3006428}, when the gradients are sparse, or very small.
Although \adagrad enjoys great benefits for sparse settings, it still suffers from the rapid decay of learning rates~\cite{reddi2018convergence}, when training nonconvex neural networks or problems with dense gradients. 
Then, the exponential moving averages of squared past gradients variants of \AdaGrad are proposed, which slow down the decay of learning rates. 
These variants, such as \Adam and \RmsProp, own better performance in practice and have become standard methods in many deep learning libraries, such as TensorFlow\footnote{https://www.tensorflow.org}, Pytorch\footnote{https://pytorch.org}. 
The \textit{Exponential Moving Average} (EMA) strategy actually gives more weights on latest (smaller) gradients.
However, recent studies~\citep{reddi2018convergence,pmlr-v70-mukkamala17a} have found that these variants may suffer from the non-convergence issues, since the aggressive EMA strategy may lead to increasing learning rates.

\AmsGrad~\citep{reddi2018convergence} is proposed to address the non-convergence issues, and uses the maximum of all exponential moving averages to avoid increasing learning rates. 
However, \AmsGrad can only be proved with a $ O( \sqrt{T }) $ regret bound, which can be much larger than the data-dependent regret bound of \AdaGrad. 
Its convergence analysis depends on the maximum of all moving averages, which can be very large and never change when choosing a bad initialization point. Thus, \AmsGrad may fall into a coordinate-wise vanilla SGD algorithm.
Another solution, \AdamNc~\citep{reddi2018convergence}, applies the equal weighting strategy on squared past gradients, which is actually a momentum based variant of \AdaGrad. Although \AdamNc achieves similar regret bound as \AdaGrad, 
it cannot give more weights on the latest (smaller) gradients as the exponential increasing weights methods.
Thus, it remains an open problem to develop new algorithms to fix the non-convergence of \Adam, which can enjoy great benefits from sparse stochastic gradients, and in the meanwhile, give more attention on recent smaller past gradients. 

To address problems mentioned above, we provide an affirmative solution in this paper. Specifically, we conclude these adaptive methods and propose the general \textit{Weighted Adaptive Gradient Methods Framework} (\WAGMF) , which include most existing adaptive methods. Based on our framework, we develop the \textit{Weighted Adaptive Algorithm} (\WADA), which not only ensures the convergence of algorithm, but also applies linear growing weights on past gradients. 
Besides, we prove that \WADA can achieve a weighted data-dependent regret bound, which can be better than the regret bound of \adagrad and may partially explain the good performance of \adam in practice. Finally, extensive experiments demonstrate the effectiveness of our methods.
The mainly contributions of this paper can be summarized as follows:

\begin{itemize}
  \item We propose a general form of several existing adaptive methods named weighted adaptive gradient method framework. Moreover, we make a detailed discussion about different weights design strategies in \WAGMF. 
  \item We develop a novel adaptive method \WADA by applying the linear growing weighting strategy to \WAGMF. We provide a convergence analysis of \WADA, and demonstrate that it can achieve a weighted data-dependent regret bound, which can be better than the data-dependent regret bound of \AdaGrad when the gradients decrease rapidly. Further, we introduce several variants of \WADA.
  \item We provide extensive experiments in comparison with existing adaptive methods on three widely used datasets. The experimental results demonstrate that our methods outperform \Adam and its variants on training convex problems and deep neural networks.
\end{itemize}

The rest of the paper is organized as follows. First, we provide a brief review of related work
in Section~\ref{sec:related}, and present some  notations and preliminaries in Section~\ref{sec:pre}. Then, we present the technical details of \WAGMF in Section~\ref{sec:wagmf}. Afterwards, in Section~\ref{sec:wada}, we implement \WADA algorithm on  \WAGMF and  give a detailed convergence analysis of \wada.
And in Section~\ref{sec:exp}, we report the performance of each algorithm on experiments.  Finally, we conclude the paper in Section~\ref{sec:con}.

\section{Related work}\label{sec:related}
In this section, we first give a brief introduction of stochastic gradients descent algorithms. Then we introduce the developments of the adaptive variants of SGD. 

SGD algorithm has attracted much attention in the field of machine learning and optimization. The update rule of SGD is to move  towards the opposite direction of stochastic gradient $ g_t = \nabla f_t(x_t) $. Many variants of SGD have been proposed and analyzed~\cite{rakhlin2012making,shamir2013stochastic,lacoste2012simpler} and have achieved great success on many problems. 
The SGD (with appropriate step sizes and averaging scheme) suffers from an  $ O (1/\sqrt{T})$ error bound for general convex problems and enjoys  an improved $ O (1/T) $ error bound for strongly convex problems.
Although these algorithms have achieved the well-known convergence rate, they may suffer from slower learning speed on sparse features, since all parameters adopt the same learning rate.

\adagrad have introduced a novel approach to utilize the historical gradients and adjust the learning rate for every parameter, which achieved great benefits from sparse settings.
Owing to its adaptive learning rates for different parameters, \adagrad has recently witnessed great potential on training deep neural networks\citep{Dean:2012wx}, where the scale of the gradients in each layer is often different by several orders of magnitude~\citep{zeiler2012adadelta}.
There are many variants of \adagrad have been developed for different tasks. For strongly convex problem, variants like \textsc{SadaGrad} \cite{Chen:2018wy}, \textsc{SC-AdaGrad} \cite{pmlr-v70-mukkamala17a}, have been proposed and also enjoyed great benefits on sparse settings.
To speed up the training process of dense or nonconvex problems, the EMA descendants of \adagrad have been developed and found to be effective for deep learning, e.g., \RmsProp~\citep{tieleman2012lecture}, \Adam~\citep{kingma2014adam}, \textsc{AdaDelta}~\citep{zeiler2012adadelta}. 
\adam not only applies the EMA technique, but also combines the momentum acceleration, which make it very popular in practice.
Then, \amsgrad and \adamnc \cite{reddi2018convergence} have been proposed  to fix the non-convergence issues  of these EMA variants. The \amsgrad have involved an additional maximization operator in \adam, which maintained the EMA technique. \adamnc have abandoned the EMA technique, and adopted the same weighting strategy as \adagrad.  
However, \amsgrad could not achieve similar regret bound as \adagrad, and \adamnc could not enjoy more benefits from the latest (smaller) gradients.

In addition, we noticed that some on going works also tried to propose new algorithms to fix the non-convergence of \adam. Nostalgic Adam \citep{Huang:2018vh} also considered using different weights on past gradients, but it only considered decreasing weighting strategy, which make it can't give more focus on the recent past gradients. New variant of \amsgrad, the \textsc{Padam}~\citep{Chen:2018tj}, also tried to fix the issues. Nevertheless, \textsc{Padam} could not be proved with $ O(\sqrt{T}) $ data-dependent regret bound as the original \amsgrad.

\section{Preliminaries}\label{sec:pre}
In this section, we first introduce some notations and assumptions. Then, we present the online convex optimization problem. Finally, we introduce the generic framework of adaptive methods and give a preliminary comparison of some existing adaptive methods.

\subsection{Notations and assumptions}
In this sequel, vectors and scalars are lower case letters, such as $ x \in \R^d $. The subdifferential set of a function $ f $ at $ x $ is denoted by $ \partial f(x) $, and the subgradient used in practice is $ g_t $,
where $ g_t \in \partial f_t(x_t)  $ and $ g_{t,j} $ denotes its $ j$-th coordinate. The $ diag(g_t) $ function changes the vector $ g_t $ to a diagonal matrix, and  $ g_{1:t} = [g_{1} ,\ g_{2} ,\ ... ,\ g_{t}] $ denotes the matrix obtained by concatenating the subgradient sequence. Furthermore, for any vector $ a \in \R^d $, we use $ \sqrt[p]{a} $ for element-wise $ p $-th root and $ a^p $ for element-wise $ p $-th power.
Let $ \mathcal{S}_d^{+} $ denote the set of all positive definite matrices in $ \R^{d \times d}$. Then, for a symmetric matrix $ V \in \mathcal{S}_d^{+} $, we denote $\sum_{i,j=1}^{d} V_{ij}x_i y_j $ as $ \langle x,Vy \rangle $ or $\langle x,y \rangle_V $, and define $ \norm{x}_V = \sqrt{ \langle x,x \rangle_V } $. 
The  weight projection $P_{\F}^V(y) $ for $ V \in \mathcal{S}_d^{+} $ on feasible set $ \F  \subset \R^d $,
is defined as $P_{\F}^V(y) = \argmin_{x \in \mathcal{F}}{ \norm{x-y}_{V}^{2} }$ for  $ y \in \R^d $ .

Similar to \citep{duchi2011adaptive,pmlr-v70-mukkamala17a}, we assume that the feasible set $ \F $ have bounded diameter i.e., $ \norm{x-y}_\infty \le D_\infty$ for every point $ x,y \in \mathcal{F} $.
Also,  we assume that the subgradient $ g_t $ has a bounded infinity norm on $ \mathcal{F} $, i.e., $ \norm{g(x_t)}_\infty \le G_\infty, \forall x_t \in \F $.

\subsection{Problem statement}
In this paper, we consider the Online Convex Optimization (OCO) problem. In the online setup, the problem is defined on a  closed convex set $ \mathcal{F} \in \R^d $. 
At each round $ t $, a loss function $ f_t : \mathcal{F} \rightarrow \R $ is then revealed, and we predict a point $ x_t \in \mathcal{F} $ with a loss $ f_t(x_t) $.
Let $ T \in \mathbb{N} $ denote the total number of iterations. The optimal goal of the problem is $ x^* = \argmin_{x \in \mathcal{F}} \sum_{t=1}^{T}  f_t(x) $. 
Then, the regret of the problem is defined as the difference between the total loss and the optimal goal. That is
\begin{equation}
R(T) = \sum_{t=1}^{T} (f_t(x_t) - f_t(x^*)) .
\end{equation}
The optimal regret bound for this problem is $ O (\sqrt{T}) $, which can be achieved by online gradient descent algorithm\citep{Zinkevich:2003tf}. The update rule of the algorithm is defined as $ x_{t+1} = P_{\F}(x_t - \alpha_t g_t) $, which moves towards the opposite direction of gradient $ g_t = \nabla f_t(x_t) $ with step size $ \alpha_t = \frac{\alpha}{\sqrt{t}} $, in the meanwhile, uses projection step to maintain next point $ x_{t+1} $ onto the feasible set $ \mathcal{F} $.  

The OCO problem is closely related to the stochastic optimization problem.
In particular, an online optimization algorithm with vanishing average regret yields a stochastic optimization algorithm for the empirical risk minimization problem \citep{cesa2004generalization,reddi2018convergence}. Thus, in this paper, we use online gradient descent and stochastic gradient descent synonymously.

\begin{algorithm}[htbp]
  \caption{Generic adaptive method}\label{alg:gen-ada}
  \begin{algorithmic}
    \STATE {\bfseries Input:} $x_1 \in \mathcal{F}$, step sizes $\{\alpha_t\}_{t=1}^{T}$,sequence of functions $ \{ \psi_t, \phi_t \}_{t=1}^T $
    \FOR { $ t=1 $ to T }
    \STATE $g_t \in  \partial  f_t(x_t)$
    \STATE $m_t = \phi_t(g_1,...,g_t) $
    \STATE $V_t = \psi_t(g_1,...,g_t) $ 
    \STATE $x_{t+1} = P_{\mathcal{F}}^{V_t} (x_t - \alpha_t V_t^{-1}  m_t ) $
    \ENDFOR
  \end{algorithmic}
\end{algorithm}

\subsection{Generic framework of adaptive method}
To demonstrate the differences of existing adaptive methods, we introduce the generic framework of adaptive methods in the work \citep{reddi2018convergence}, which includes most existing adaptive methods. As the generic adaptive method is shown in Algorithm~\ref{alg:gen-ada}, the main difference between adaptive methods is the choice of the "averaging"  function $\phi_t $ and  $ \psi_t $. Here $ \phi_t: \F^t \rightarrow \R^d $ , $ \psi_t:  \F^t \rightarrow \mathcal{S}_d^+ $ and $ V^{-1} $ represents the inverse of matrix $V$. Different from its original form in  \citep{reddi2018convergence}, we consider $ V_t $ as a generic adaptive learning rate estimation matrix, not limited to the square root form.

Generally speaking, the averaging function $\phi_t$  is to get an approximation of the real gradient $ \frac{1}{T} \sum_{i}^{T} \partial f_i(x_t)  $.
Since each function only comes at one round, we have $ E( g_t ) =  \frac{1}{T} \sum_{i}^{T} \partial f_i(x_t) $. 
Therefore, $ g_t $ is a good approximation of the real gradient. 
However, the momentum form of gradient $ g_t $ is more popular in practice which appears to significantly boost the performance~\citep{reddi2018convergence}. 
The averaging function $ \psi_t $ is an approximation of the inverse matrix of adaptive learning rates. 
$ \psi_t $ is the key secret of adaptive methods. 
We demonstrate the averaging function $ \phi_t $  and $\psi_t $  of several adaptive methods in the Table~\ref{tab:ada-alg}. 

\begin{table}[htbp]
  \centering
  \caption{The averaging function  $\phi_t$, $\psi_t$  of different adaptive methods}\label{tab:ada-alg}
  \begin{footnotesize}
    \renewcommand\arraystretch{1.8}
    \begin{tabular*}{\columnwidth}{p{0.3\columnwidth}p{0.26\columnwidth}p{0.4\columnwidth}}
      \toprule 
      \specialrule{0em}{-1pt}{-1pt}
      Algorithm & $\phi_t$ & $\psi_t$  \\ 
      \midrule
      Online GD& $ g_t $ & $I$ \\
      \textsc{signSGD}\footnote{We refer \citep{Bernstein:2018wn} for this method.} & $ g_t $ & $ diag(\norm{ g_t} )  $ \\
      \AdaGrad &  $ g_t $ & $ diag( \sqrt{\sum_{i}^{t} g_i^2 } )$  \\ 
      \RmsProp &  $ g_t $ & $ diag( \sqrt{\sum_{i=1}^{t} \beta_2^{t-i+1} g_i^2} ) $   \\ 
      \RmsProp\footnote{The \RmsProp's revision in \citep{pmlr-v70-mukkamala17a}. we use $ \beta_t = 1 - 1/t $ .} &  $ g_t $&  $  diag( \sqrt{\sum_{i}^{t} g_i^2 /t } )$ \\ 
      \Adam\footnote{For \Adam and its variants, we omit the bias correction terms.}   &  $ \sum_{i=1}^{t} \beta_1^{t-i+1} g_i $ &  $ diag( \sqrt{\sum_{i=1}^{t} \beta_2^{t-i+1} g_i^2} ) $ \\ 
      \AdamNc  & $ \sum_{i=1}^{t} \beta_1^{t-i+1} g_i$ & $ diag ( \sqrt{\sum_{i}^{t} g_i^2 /t } )$  \\ 
      \AmsGrad &  $ \sum_{i=1}^{t} \beta_1^{t-i+1} g_i $ & $ \sqrt{ \max(\psi_t^2, (\psi_t^2)^{adam} ) }  $ \\ 
      \bottomrule
    \end{tabular*} 
  \end{footnotesize}
\end{table}

\section{Weighted adaptive gradient method framework} \label{sec:wagmf}
In this section, we first review some existing adaptive methods in weighted past gradient perspective. Then, we formally introduce the technical details of the \textit{Weighted Adaptive Gradient Method Framework} (\WAGMF). At last, we  make a detailed discussion about three weighting strategies in \WAGMF. 

\subsection{Adaptive methods revisit}

Before introducing our framework, we revisit the averaging function $ \psi_t $ of some existing adaptive methods. 
The pioneer of adaptive method, \AdaGrad, sums all squared gradient equally in in the averaging function. The formal averaging function $ \psi_t^{adagrad} $ is:
\begin{equation}
\frac{1}{t}\psi_t^2(g_1,...,g_t) = \frac{ diag( \sum_{i=1}^{t}  g_i^2 ) }{t} .
\end{equation}
These variants of \adagrad, such as \Adam and \rmsprop, use the exponential moving averages of squared past gradients in the averaging function $ \psi_t $.
If we set \Adam with a constant $ \beta_2  \in [0,1) $, the averaging function $ \psi_t^{adam} $ is:
\begin{equation}
\frac{1}{t} \psi_t^2(g_1,...,g_t) =  (1-\beta_2) diag(  \sum_{i=1}^{t} \beta_2^{t-i} g_i^2 ) .
\end{equation}

By comparing the two averaging functions, we can find that the main difference is that \Adam uses exponential growing weights on squared past gradients while \AdaGrad utilizes equal weights. 
Intuitively, the recent past gradients are more accurate approximations of the current gradient than the early past ones, since the recent points are generally close to the current point. 
Based on this intuition and the good performance of \Adam in practice, we think using growing weights on past gradients should be reasonable in adaptive methods.

\begin{algorithm}[t]
  \caption{Weighted adaptive gradient methods framework}\label{alg:wagmf}
  \begin{algorithmic}
    \STATE {\bfseries Input:} $x_1 \in \mathcal{F}$, step sizes $\{\alpha_t\}_{t=1}^{T}$,$\{\beta_{1t} \}_{t=1}^T$, $\{\gamma_{t} \}_{t=1}^T$, gradient power $ p_1 $,  the $ p_2$-th root
    \STATE {\bfseries Initialize: set $m_0 = 0, v_0 = 0$ }
    \FOR { $ t=1 $ to T }
    \STATE $g_t \in  \partial  f_t(x_t)$
    \STATE $m_t = \beta_{1t} m_{t-1} + (1-\beta_{1t})g_t $
    \STATE $v_t = v_{t-1} +  \gamma_t  g_t^{p_1} $ 
    \STATE $ b_t = \frac{ 1 }{ \sum_{i=1}^{t} \gamma_i }  $, $V_t = \diag{\sqrt[p_2]{ v_t \cdot  b_t } }$
    \STATE $x_{t+1} = P_{\mathcal{F}}^{V_t} (x_t - \alpha_t V_t^{-1}  m_t ) $
    \ENDFOR
  \end{algorithmic}
\end{algorithm}

\subsection{Details of \WAGMF}
Since the key difference is the weighting strategy, we can choose different weighting strategies to design new adaptive methods. Here, we propose the \textit{weighted adaptive gradient method framework} which attaches different weights on the past gradients.
The formal algorithm is presented in Algorithm~\ref{alg:wagmf}.
Specifically, for the averaging function $ \phi_t $,  \wagmf chooses the most widely used momentum form as it appears to significantly boost the performance~\citep{reddi2018convergence}.
And for the key part of the averaging function $ \psi_t $, \wagmf use the $ p_1 $-th power of gradients instead of the squared gradients, which have been adopted by most existing adaptive methods. 
What's more, \wagmf attaches weight $ \gamma_t $ on the past gradients $ g_t^{p_1} $. 
Since all past gradients are in the weighted form, \wagmf introduce another weight balance term $ b_t =  \frac{ 1 }{ \sum_{i=1}^{t} \gamma_i } $ to get the weighted average of all gradients. 
To correspond with $ p_1 $-th power of gradients, \wagmf adopt $ p_2 $-th root to get the final estimation matrix $ V_t $ instead of the squared root.
There are more details in Algorithm~\ref{alg:wagmf}.

To fix the non-convergence issues of \adam, we should add additional condition to avoid increasing learning rate. For any adaptive methods based on \wagmf, we should ensure that $ \frac{ b_t^{-p_2} }{\alpha_t} \ge \frac{ b_{t-1}^{-p_2}} {\alpha_{t-1}} $ for any $ t $.
By adding this condition, we follow the convergence analysis in  \citep{reddi2018convergence,kingma2014adam}, and present the key result for \WAGMF in Theorem~\ref{thm:conv}. More details of the proof are presented in appendixes.

\begin{theorem}\label{thm:conv}
  Let $ \{x_t \} $  be the sequence generated by \WAGMF (Algorithm~\ref{alg:wagmf}). Assume that $ \mathcal{F} $ has bounded diameter $ D_\infty $, the subgradient $ g_t $ has bounded infinity norm $ G_\infty $  and $ \frac{ b_t^{-p_2} }{\alpha_t} \ge \frac{ b_{t-1}^{-p_2}} {\alpha_{t-1}} $, then for any $ x^* \in \mathcal{F} $, \WAGMF have following regret bound:
  \begin{align*}
  R(T) \le & 
        \frac{D_\infty^2}{2 \alpha_T (1-\beta_{1})} \sum_{i=1}^{d}  V_{T,i} 
      + \frac{D_\infty^2}{2 } \sum_{t=1}^{T} \sum_{i=1}^{d}
      \frac{\beta_{1t}  V_{t-1,i} }{ (1-\beta_{1t}) \alpha_t }
      \\& + \sum_{t=1}^{T}  \frac{\alpha_t}{1-\beta_1} \norm{m_t}_{V_t^{-1}}^2 .
  \end{align*}
\end{theorem}

Theorem~\ref{thm:conv} gives a basic convergence result for \WAGMF. The regret bound is mainly construct by three terms which can be further bounded by setting other hyper-parameters. So, we can design ideal adaptive methods by implementing different weighting strategies on \WAGMF.

\subsection{Weighting strategy}
Weighting strategy is the critical component of \WAGMF, and a suitable weighting strategy plays a key role in the adaptive method. 
Here, we make a discussion about three types of weights design strategies, i.e., equal weights, increasing weights and decreasing weights.

\textbf{Equal weights:}
Equal weights maybe the simplest and most useful strategy. \AdaGrad and \AdamNc use this strategy and achieve the data-dependent regret bound. So far, to the best of our knowledge, equal weighting strategy is the only strategy that can achieve the well-known data-dependent regret bound $ O(\sum_{i=1}^{d} \norm{ g_{1:T,i} }_2 ) $. In this paper, we extend the original squared gradients in $ \psi_t $ to the $ p $-th power of gradients and set $ p_1 = p_2 = p $.  Applying the conclusion in Theorem~\ref{thm:conv}, we have following regret bound for this situation:

\begin{theorem}\label{thm:equ-w}
  Let $ \{x_t \} $  be the sequence generated \WAGMF in Algorithm~\ref{alg:wagmf}. Assume all conditions are held in Theorem~\ref{thm:conv}, we set equal weights $ \gamma_t = 1 $ on past gradients and set $p_1 = p_2 = p  $  where $  p = 2^s (s \ge 1, s \in N_{+}) $, then for any $ x^* \in \mathcal{F} $, we have the following regret bound:
  \begin{align*}
   R(T) &\le \frac{D_\infty^2}{2 (1-\beta_{1})} T^{1/2 -1/p} \sum_{i=1}^{d} \norm{g_{1:T,i}}_p  
  \\& + \frac{ \beta_1 D_\infty^2 G_\infty  }{2 (1-\beta_{1}) (1-\lambda)^2 } 
   + \frac{2 \alpha}{(1-\beta_1)^3}   \sum_{i=1}^{d} \norm{g_{1:T,i} }_2 .
  \end{align*}
\end{theorem}

Let $ p=2 $, the above theorem can get the data-dependent regret bound $ O(\sum_{i=1}^{d} \norm{ g_{1:T,i} }_2 ) $. For the worst case, Theorem~\ref{thm:equ-w} can achieve the $ O(\sqrt{T}) $ regret bound. Since the gradients values are involved in the bound terms, the regret bound can get significant improvement for sparse or small gradients.

\textbf{Increasing weights:}
The EMA variants adaptive methods use exponential increasing weights on squared past gradients. \RmsProp, \Adam, \textsc{Nadam} \citep{dozat2016incorporating}, and \textsc{AdaDelta} are some distinguished algorithms which fall in this category. 
The weight sequence $ \gamma_t $ of them is set as:
\begin{equation}
 \gamma_t = \frac{1}{\beta_2^t},
\end{equation}
where $ 0 < \beta_2 < 1 $.  Recently proposed \textsc{signSGD} \citep{Bernstein:2018wn} adopts a more extreme increasing weighting strategy:
\begin{equation}
\gamma_{t} = \infty \cdot \gamma_{t-1} ,
\end{equation}
where the next weight $ \gamma_{t} $ is the higher order infinity of $ \gamma_{t-1} $.
These aggressive weighting strategies may lead to the increasing learning rates, which make these methods can not converge for general OCO problem.
Although there are some convergence issues for these methods, they are very popular and have made a significant contribution for the deep learning community. 

\textbf{Decreasing weights:}
Decreasing weighting strategy is not widely used in practice. Actually, if the weight $ \gamma_{t} $ gradually decays to zero, the latest (smaller) gradient is less important, so we can hardly obtain benefits from the latest sparse or small gradients.
Nostalgic Adam falls in the decreasing weighting strategy category which \cite{Huang:2018vh} consider following hyper-harmonic series as the weight sequences:
\[ \gamma_t = \frac{1}{ t^{\eta}}, \eta \ge 0.\]
If we choose $ \eta > 0 $, $ \gamma_{t} $ will decay to $ 0 $ over time, and the estimation matrix $ V_t $ will hardly change. So, these methods may behave like the momentum based vanilla SGD algorithm.

\section{Weighted adaptive algorithm}\label{sec:wada}
In this section, we introduce a novel \textit{weighted adaptive algorithm} (\WADA) to fix the non-convergence issues of \Adam, which implements a linear growing weighting strategy on  \WAGMF. \WADA is more consistent with the original \Adam in the perspective of weighting strategy. Besides, we also propose some variants of \WADA.

\subsection{Details of \WADA}
Because of the aggressive exponential increasing weighting strategy, the EMA adaptive methods like \Adam, may lead to the non-convergence issues~\citep{reddi2018convergence}. Since the \AdamNc chooses an equal weighting strategy to fix the issues, 
there may still be some doubts whether increasing weights strategy would lead to the non-convergence, and whether it is possible to design new adaptive methods, which adopts increasing weighting strategy and guarantees the convergence.
To answer these problems, we consider applying a milder increasing weighting strategy to \WAGMF, the simplest linear growing weighting strategy. 
We propose the \textit{Weighted Adaptive Algorithm} (\WADA) based on this idea which attaches linear growing weights on squared past gradients. The formal algorithm of \wada is presented in Algorithm~\ref{alg:wada}.  

\begin{algorithm}[htbp]
  \caption{Weighted adaptive algorithm (\WADA)}\label{alg:wada}
  \begin{algorithmic}
    \STATE {\bfseries Input:} $x_1 \in \mathcal{F}$, step sizes $\{\alpha_t\}_{t=1}^{T}$,$\{\beta_{1t} \}_{t=1}^T$, 
    \STATE {\bfseries Initialize: set $m_0 = 0, v_0 = 0$ }
    \FOR { $ t=1 $ to T }
    \STATE $g_t \in  \partial  f_t(x_t)$
    \STATE $m_t = \beta_{1t} m_{t-1} + (1-\beta_{1t})g_t $
    \STATE $v_t = v_{t-1} +  t \cdot g_t^{2} $ 
    \STATE $V_t = \diag{\sqrt[4]{ \frac{2 v_t}{t(t+1)}} }$
    \STATE $x_{t+1} = P_{\mathcal{F}}^{V_t} (x_t - \alpha_t V_t^{-1}  m_t ) $
    \ENDFOR
  \end{algorithmic}
\end{algorithm}

\WADA uses linear growing weights sequence $\{ \gamma_{t} = t \}  $ and set $ p_1 = 2 $ to get the squared gradients.
The linear growing weights lead the $ \sum_{i=1}^{t} \gamma_t = O(t^2) $ ,  so we set $ p_2 = 4 $ to ensure the assumption $ \frac{ b_t^{-p_2} }{\alpha_t} \ge \frac{ b_{t-1}^{-p_2}} {\alpha_{t-1}} $ in Theorem~\ref{thm:conv}.
For the convergence analysis of \wada, we present following key result:

\begin{theorem}\label{thm:main}
  Let $ \{x_t \} $  be the sequence generated \WADA in Algorithm~\ref{alg:wada}. Assume all conditions are held in Theorem~\ref{thm:conv}, $ \gamma_{t} = t, p_1 = 2 , p_2 = 4$ and $ \alpha_{t} = \frac{\alpha}{\sqrt{t}} $, for any $ x^* \in \mathcal{F} $, we have the following regret bound:
  \begin{align*}
  \small
   & R(T)  \le 
   \frac{D_\infty^2}{2 (1-\beta_{1})} \sum_{i=1}^{d} \sqrt[4]{\sum_{j=1}^{T} j \cdot g_{j,i}^2}
   \\& + \frac{D_\infty^2}{2 } \sum_{t=1}^{T} \sum_{i=1}^{d}
  \frac{\beta_{1t}  V_{t-1,i} }{ (1-\beta_{1t}) \alpha_t }
  + \frac{\alpha d G_\infty}{(1-\beta_{1})^2} \sum_{i=1}^{d} \sqrt[4]{\sum_{j=1}^{T} j  \cdot g_{j,i}^2} .
  \end{align*}
\end{theorem}

By setting the $ \beta_{1t} $ sequence, we can get following corollary for Theorem~\ref{thm:main}.

\begin{corollary}
  Suppose $ \beta_{1t} = \beta_{1} \cdot \lambda^{t-1}$ in Theorem~\ref{thm:main}, since $ V_{t-1,i} \le \sqrt{ G_\infty}$, we have
  \begin{align*}
  R(T)  & \le 
  \frac{D_\infty^2}{2 (1-\beta_{1})} \sum_{i=1}^{d} \sqrt[4]{\sum_{j=1}^{T} j \cdot  g_{j,i}^2}
   +  \frac{\beta_{1} D_\infty^2 \sqrt{G_\infty} }{2(1-\beta_{1}) (1-\lambda)^2 } 
  \\& + \frac{\alpha d G_\infty}{(1-\beta_{1})^2} \sum_{i=1}^{d} \sqrt[4]{\sum_{j=1}^{T} j \cdot  g_{j,i}^2} .
  \end{align*}
\end{corollary}
The above corollary shows that \wada achieves an $ O(\sum_{j=1}^{d} \sqrt[4]{\sum_{i=1}^{T}i \cdot g_{i,j}^2}) $ weighted data-dependent regret bound. 
Similar to the data-dependent regret bound  $ O( \sum_{j=1}^{d} \sqrt{ \sum_{i=1}^{T} g_{i,j}^2 ) } $ of \adagrad \citep{duchi2011adaptive}, the above bound can be considerably smaller than the $ O(\sqrt{T}) $ regret bound. 
When the gradients are sparse or in general small, most of values $ g_{i,j} $ are close to zero, which makes the above bound $ \sum_{j=1}^{d} \sqrt[4]{\sum_{i=1}^{T}i \cdot g_{i,j}^2} \ll \sqrt{T} $. 
The intuition of \WADA is to give more focus on the recent past gradients, and the above weighted data-dependent regret bound also shows that the latter gradients have more influence on the bound. For a general optimization task, the gradients at first can be very large because of bad initialization point, while the gradients will become very small when it comes close to optimal point (or critical point). As weighted data-dependent regret bound is well consistent with the feature of optimization task, we think weighted data-dependent regret bound can be better than the well-known bound of \adagrad when the gradients decrease rapidly.

Although many adaptive methods have been using increasing weighting strategy on squared past gradients, none of them could be proved with a weighted form data-dependent regret bound, not even data-dependent regret bound. 
The weighted data-dependent regret bound shows that we can indeed obtain more benefits from recent past small gradient, by applying increasing weights on  past gradients. This may partially explain the good performance of other adaptive methods  (such \adam and \rmsprop) with increasing weights strategy.

\subsection{Variants of \WADA}
\label{sec:extensions}
In this subsection, we present several variants of \WADA. We consider a more general form of \WADA, which uses different the $ p_1 $-th power in the averaging function $ \psi_t $. 
Specifically, among most of existing adaptive methods, the result of $ \phi_t $ and $ \psi_t $ have the same order compared with $ g_t $. 
In the design of the adaptive methods, if we take  $ g_t $ as a random variable (to be precise, not), then it is held for most of adaptive methods,
\begin{equation}\label{equ:ada-priciple}
\phi_t =  \Theta (g_t) , \psi_t = \Theta( diag\{ g_t\}) .
\end{equation}
Following Equation~\ref{equ:ada-priciple}, we have
\begin{equation}
 \frac{\phi_t}{\psi_t} \approx \frac{\Theta (g_t)}{\Theta(diag\{ g_t\})}  \approx sign(g_t) .
\end{equation}
But for \WADA, we use $ \psi_t = \Theta( diag\{\sqrt{ g_t}\}) $ instead of $  \Theta(diag\{ g_t\}) $ , which is the biggest difference between existing adaptive methods and \WADA. Therefore, we extend the original \WADA by changing the hyper-parameter $ p_1 $ to get different averaging function $ \psi_t $. In this paper, we introduce two variants, \threeada ($ p_1 = 3 $) and \fourada ($ p_1 = 4$). 

Due to the fact that $ v_t $ can be very large in \WADA and its variants, we introduce another numerically stable iterative form for these methods. Let $ v_t^{new} = v_t* b_t $, since $ \gamma_t=t \ $ and $  \ p_2=4 $, we can use the $ v_t^{new} $ in the algorithm instead of $ v_t $:
\begin{equation}
\begin{aligned}
&v_t = v_{t-1} +  t \cdot  g_t^{p_1} \\
&\Rightarrow v_t^{new} \cdot \frac{1}{b_t} = v_{t-1}^{new} \frac{1}{b_{t-1}}  +t \cdot g_t^{p_1} \\
&\Rightarrow v_t^{new} \frac{t(t+1)}{2} = v_{t-1}^{new} \frac{t(t-1)}{2} + t \cdot g_t^{p_1} \\
&\Rightarrow v_t^{new} = (1- \frac{2}{t+1})  v_{t-1}^{new} +  \frac{2}{t+1}  g_t^{p_1} .
\end{aligned}
\end{equation}

\begin{algorithm}[htbp]
  \caption{Numerically stable iterative form of \WADA($ p $=2) , \threeada($ p $=3) and  \fourada($ p $=4) }
  \label{alg:variants-ada}
  \begin{algorithmic}
    \STATE {\bfseries Input:} $x_1 \in \mathcal{F}$, step sizes $\{\alpha_t\}_{t=1}^{T}$,$\{\beta_{1t} \}_{t=1}^T$, $\{\gamma_{t} \}_{t=1}^T$, gradient order $ p $
    \STATE {\bfseries Initialize: set $m_0 = 0, v_0 = 0$ }
    \FOR   { $ t=1 $ to T }
    \STATE $g_t \in  \partial  f_t(x_t)$
    \STATE $m_t = \beta_{1t} m_{t-1} + (1-\beta_{1t})g_t $
    \STATE $v_t = (1- \frac{2}{t+1})  v_{t-1} +  \frac{2}{t+1}  g_t^p  $ 
    \STATE $V_t = \diag{\sqrt[4]{ v_t} }$, 
    \STATE $x_{t+1} = P_{\mathcal{F}}^{V_t} (x_t - \alpha_t V_t^{-1}  m_t ) $
    \ENDFOR
  \end{algorithmic}
\end{algorithm}

Given the new update rule, we present the numerically stable version of \wada in Algorithm~\ref{alg:variants-ada}. Similar to \adam~\citep{kingma2014adam}, \wada and its variants are also easy to implement and have high computational efficiency and low memory requirements, which making them ideal for problems that are large in terms of data and/or parameters.
All experiments for our methods are based on the numerically stable iterative form.

\section{Experiments} \label{sec:exp}

\begin{figure*}[ht] 
  \centering
  \subfigure{
    \includegraphics[width=0.32\linewidth]{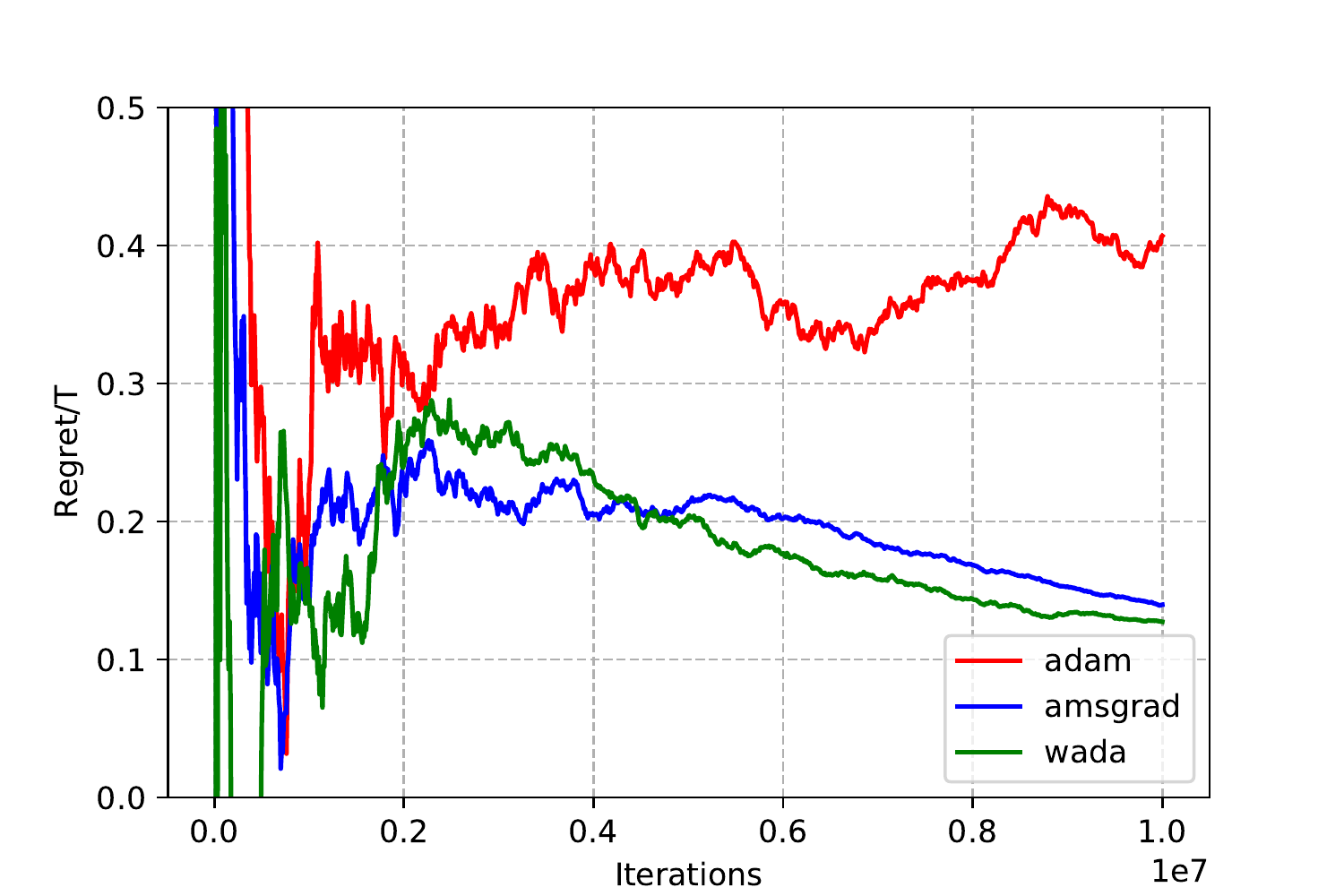}
  }
  \subfigure{
    \includegraphics[width=0.32\linewidth]{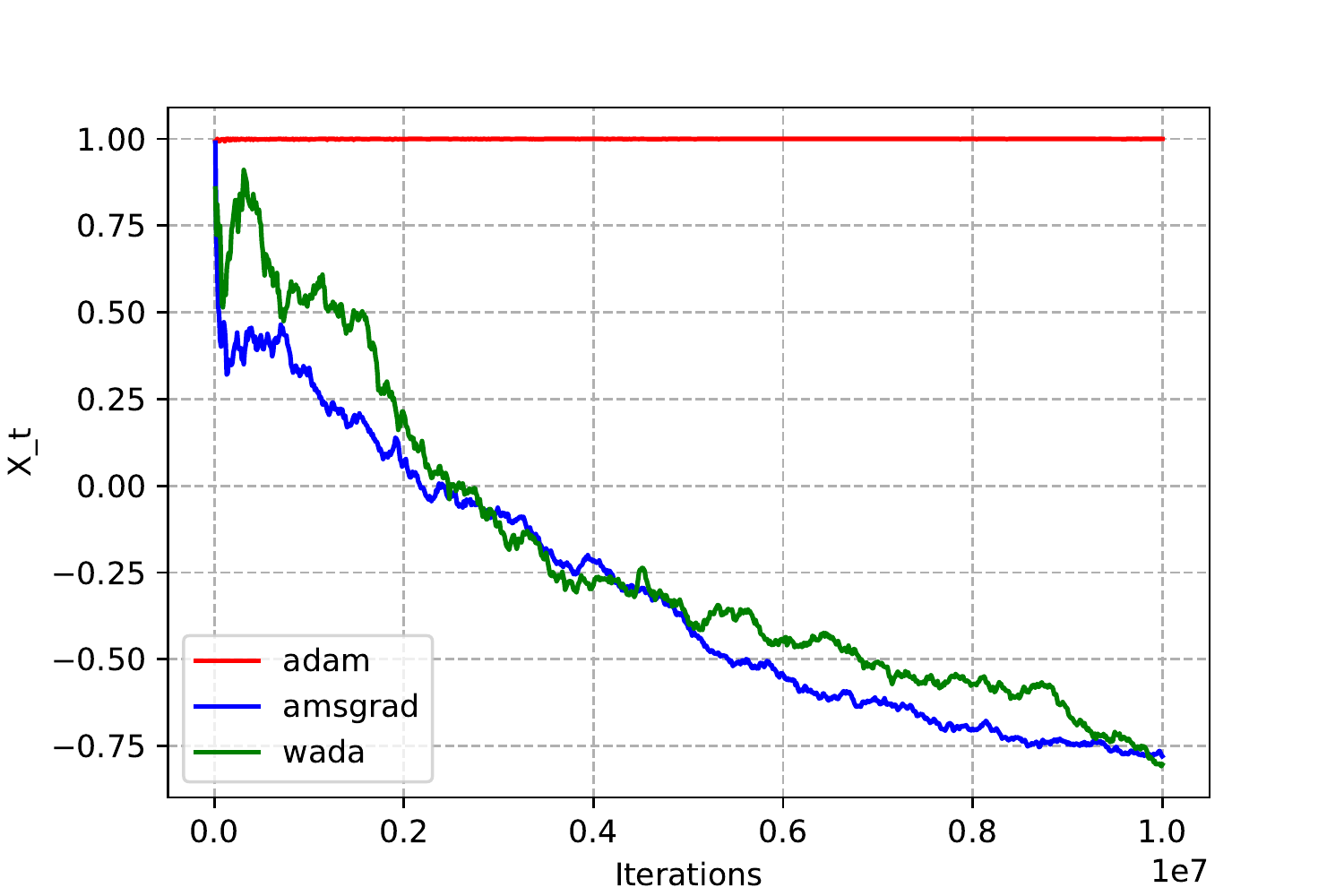}
  }
  \subfigure{
    \includegraphics[width=0.32\linewidth]{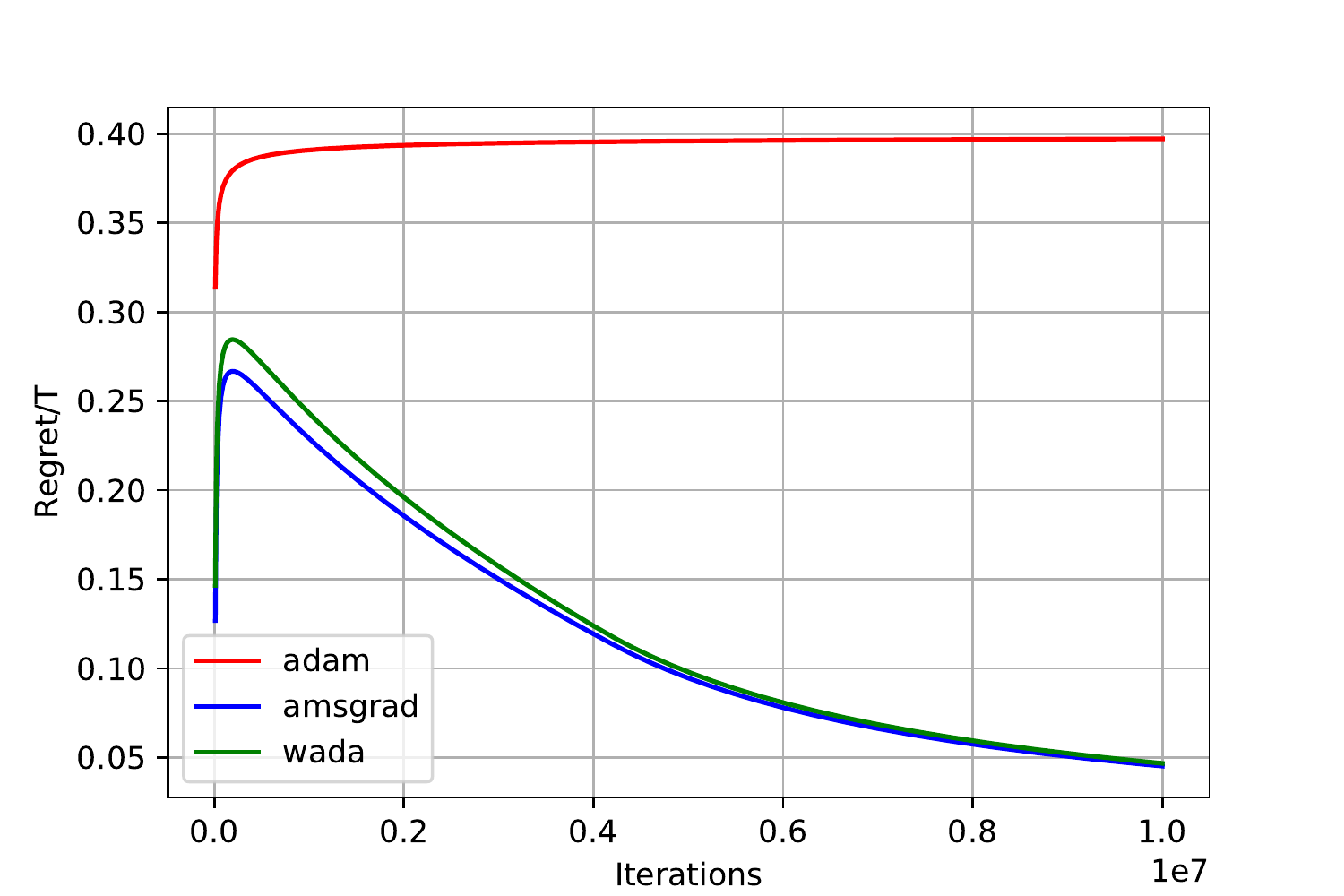}
  }
  \caption{Synthetic Experiments Results. Performance comparison of \adam, \amsgrad and \wada on synthetic experiments. The first two plots (left and center) are $ R(t)/t $ and $ x_t $ vs iteration numbers for stochastic setting. For comparison, we also show the $ R(t)/t $ plot (right) for online setting.}
  \label{fig:synthetic}
\end{figure*}

\begin{figure*}[ht] 
  \centering
  \subfigure[MNIST]{
    \includegraphics[width=0.32\linewidth]{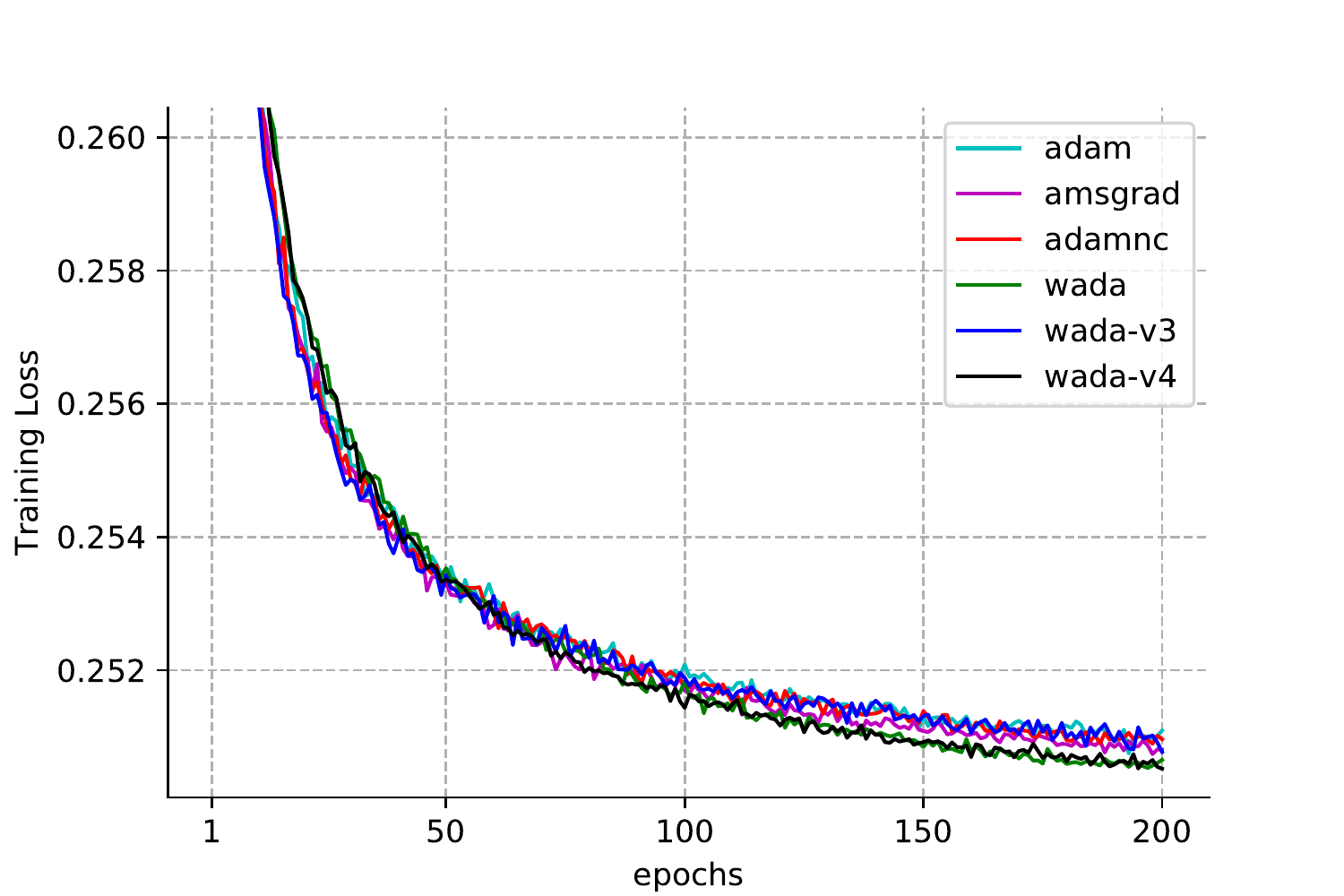}
  }
  \subfigure[CIFAR10]{
    \includegraphics[width=0.32\linewidth]{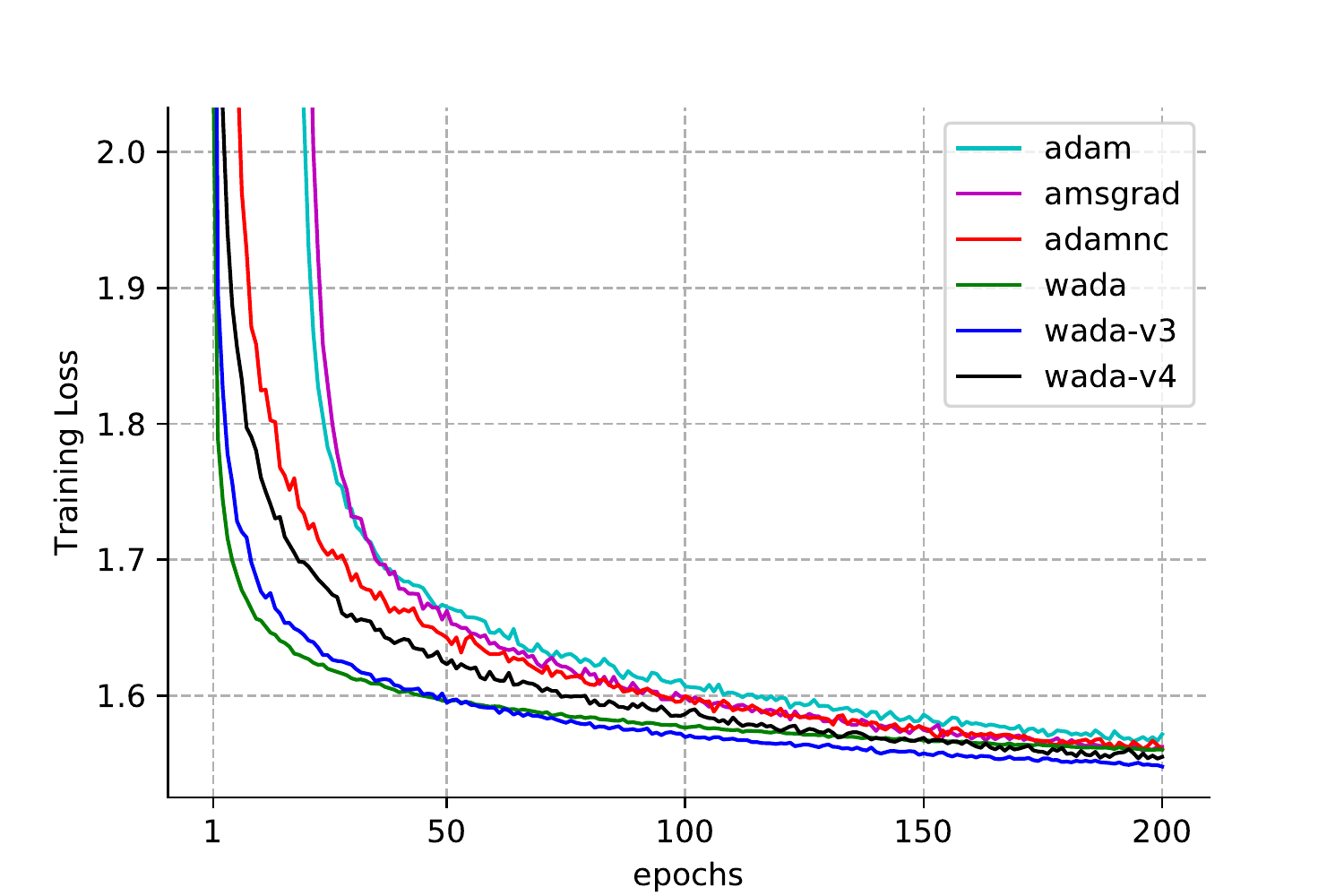}
  }
  \subfigure[CIFAR100]{
    \includegraphics[width=0.32\linewidth]{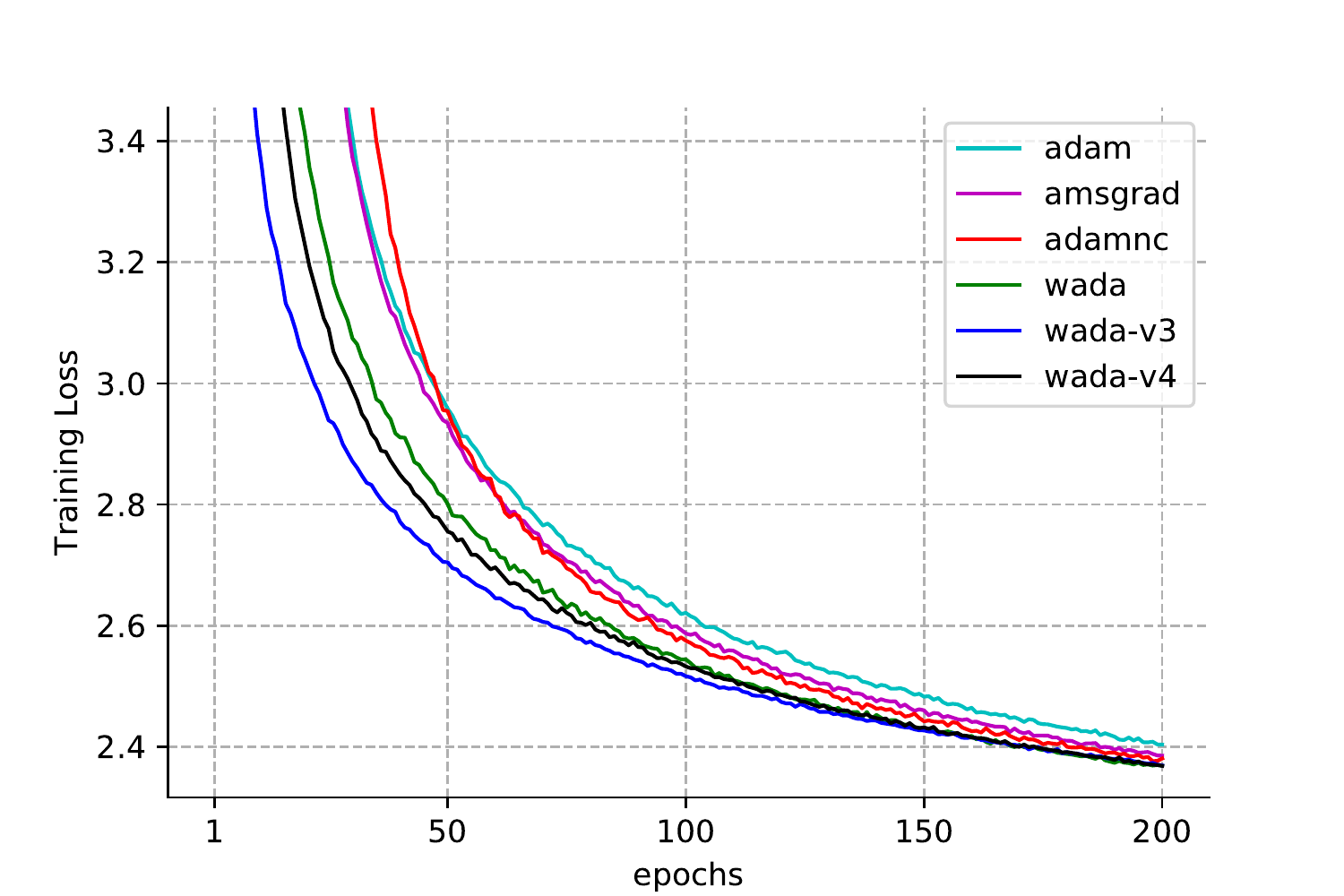}
  }
  \caption{Training Loss vs Number of Epoch for Softmax Regression on MNIST, CIFAR10 and CIFAR100 dataset.}
  \label{fig:loss-lr}
\end{figure*}

\begin{figure*}[ht] 
  \centering
  \subfigure[MNIST]{
    \includegraphics[width=0.32\linewidth]{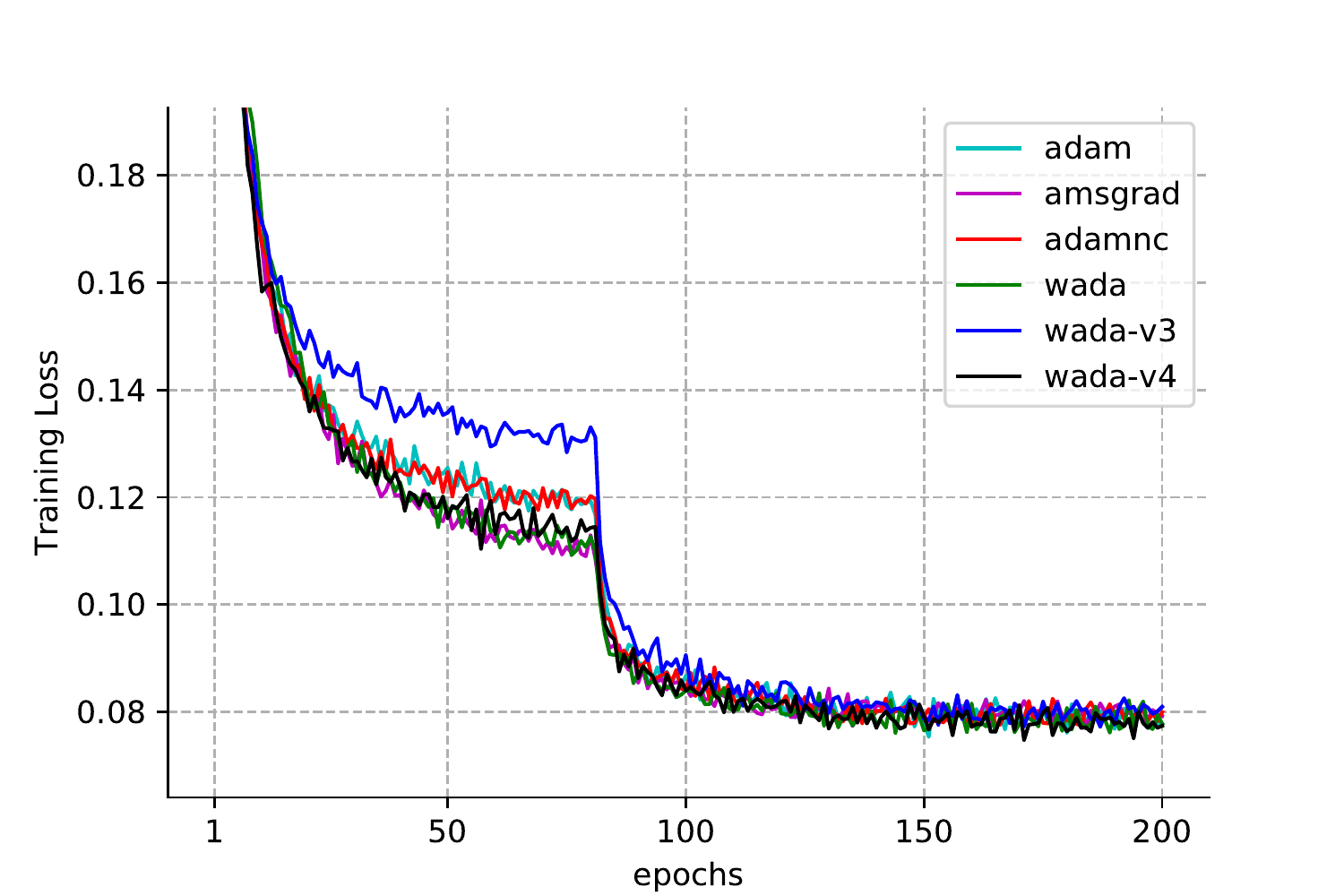}
  }
  \subfigure[CIFAR10]{
    \includegraphics[width=0.32\linewidth]{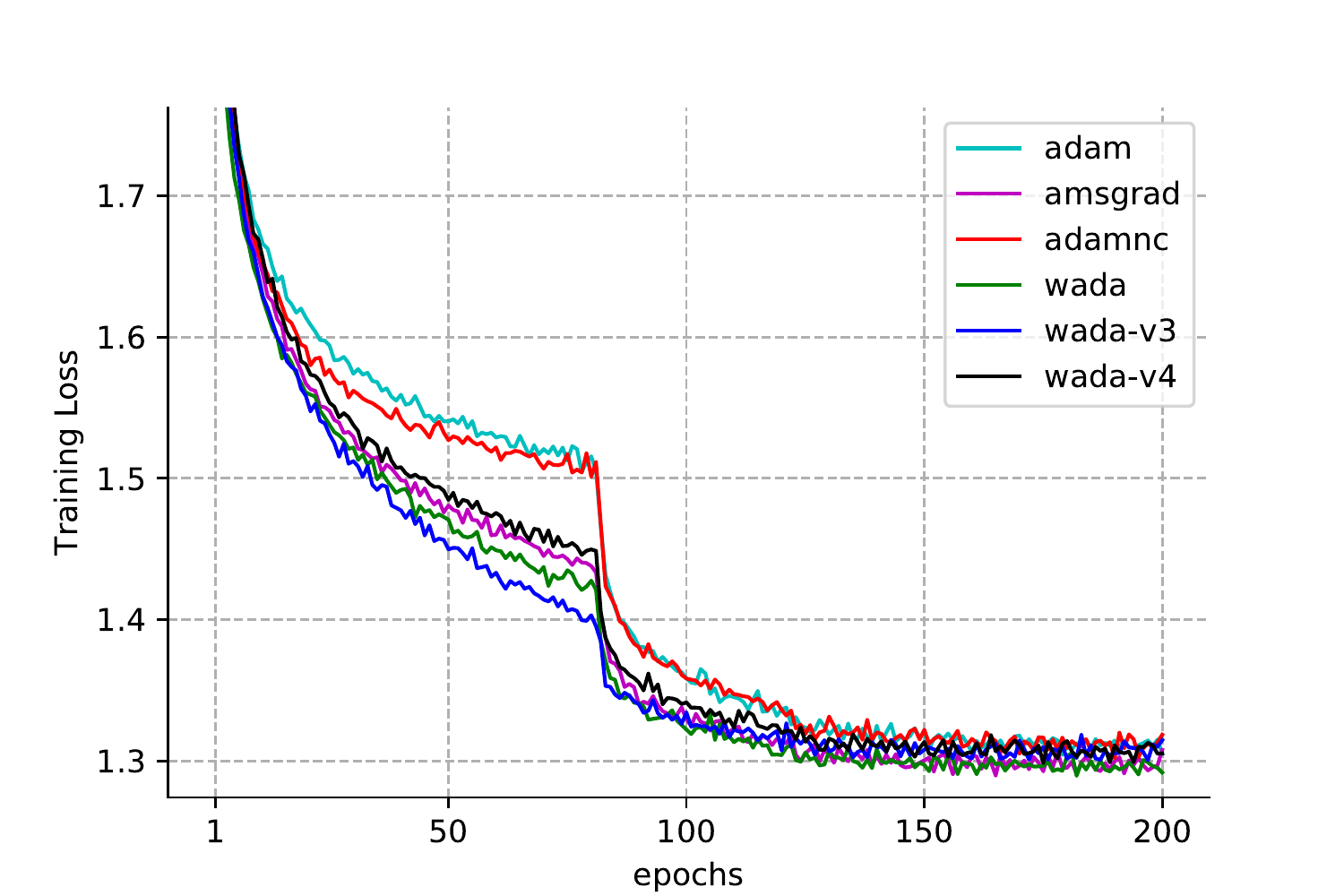}
  }
  \subfigure[CIFAR100]{
    \includegraphics[width=0.32\linewidth]{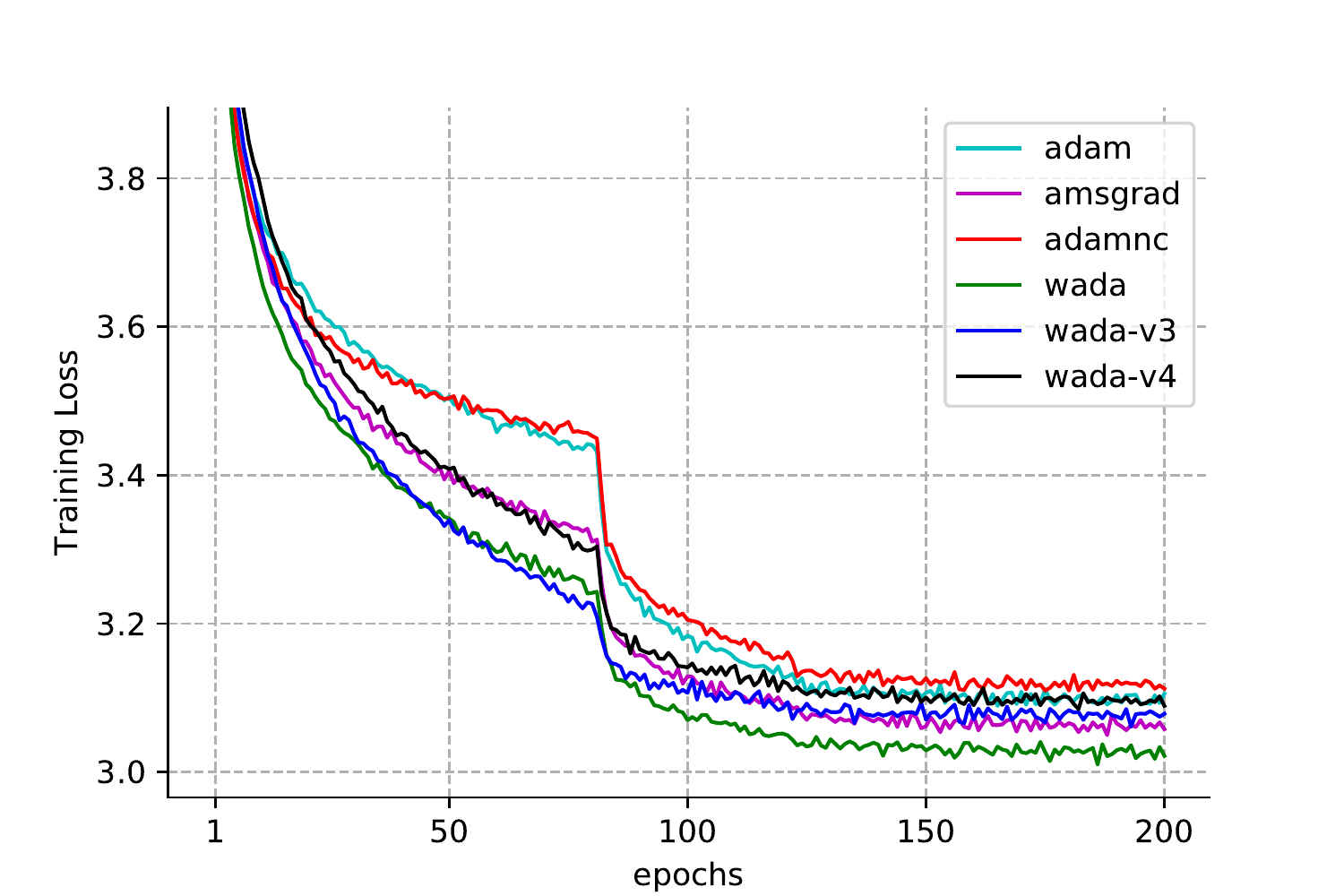}
  }
  \caption{Training Loss vs Number of Epoch for 3-layer MLP on MNIST, CIFAR10 and CIFAR100 dataset.}
  \label{fig:loss-mlp}
\end{figure*}

\begin{figure*}[ht] 
  \centering
  \vspace{0em}
  \subfigure[MNIST]{
    \includegraphics[width=0.32\linewidth]{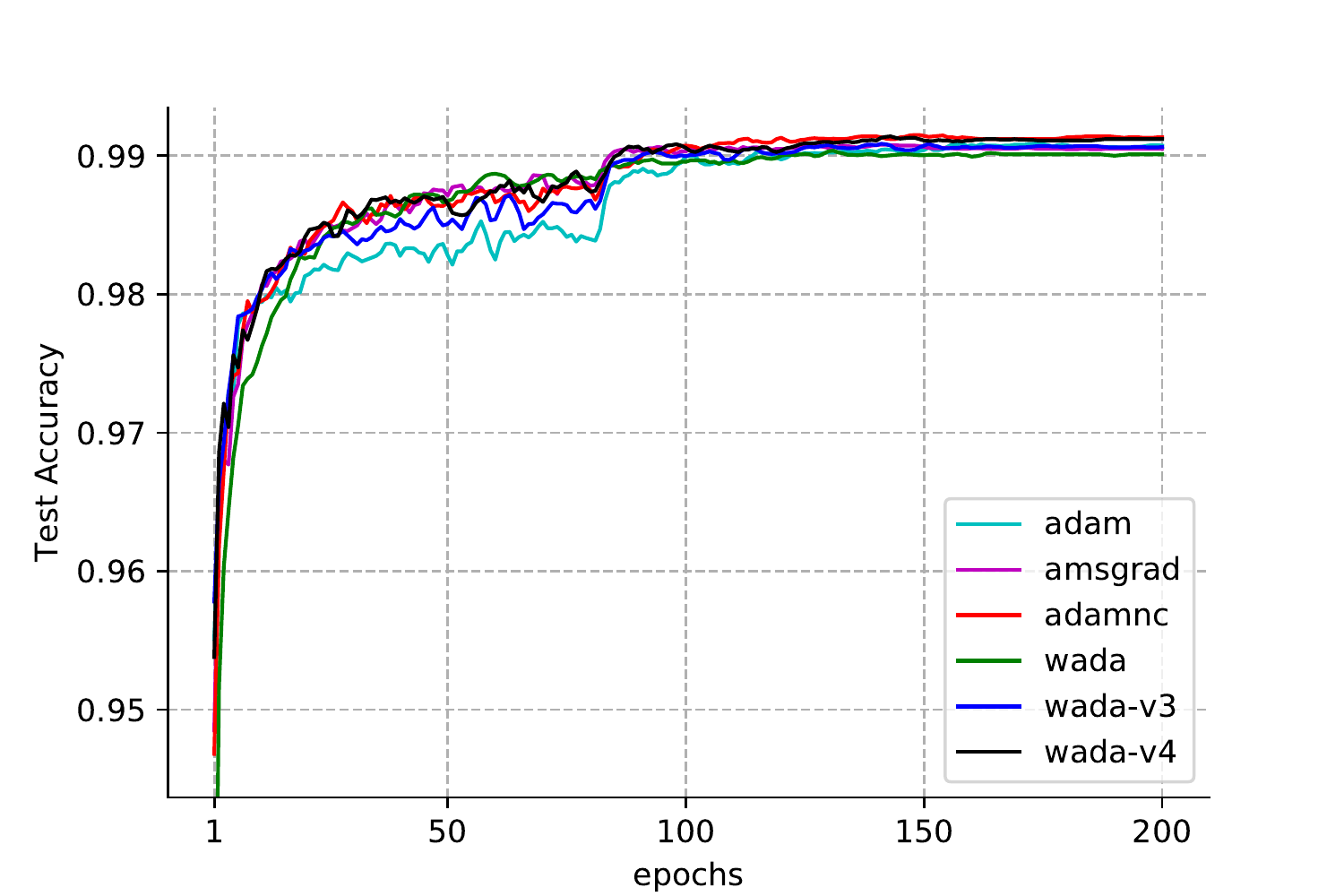}
  }
  \subfigure[CIFAR10]{
    \includegraphics[width=0.32\linewidth]{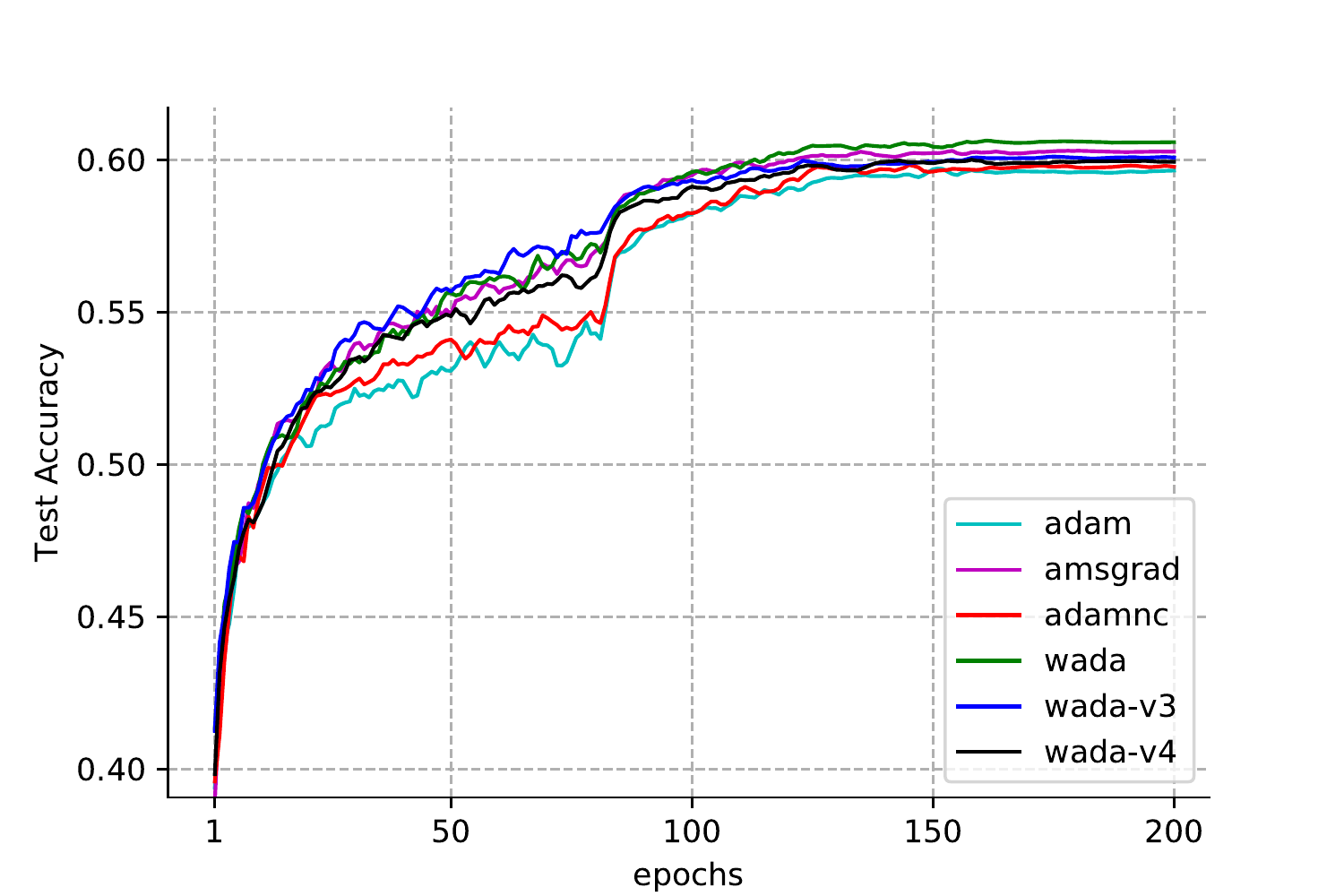}
  }
  \subfigure[CIFAR100]{
    \includegraphics[width=0.32\linewidth]{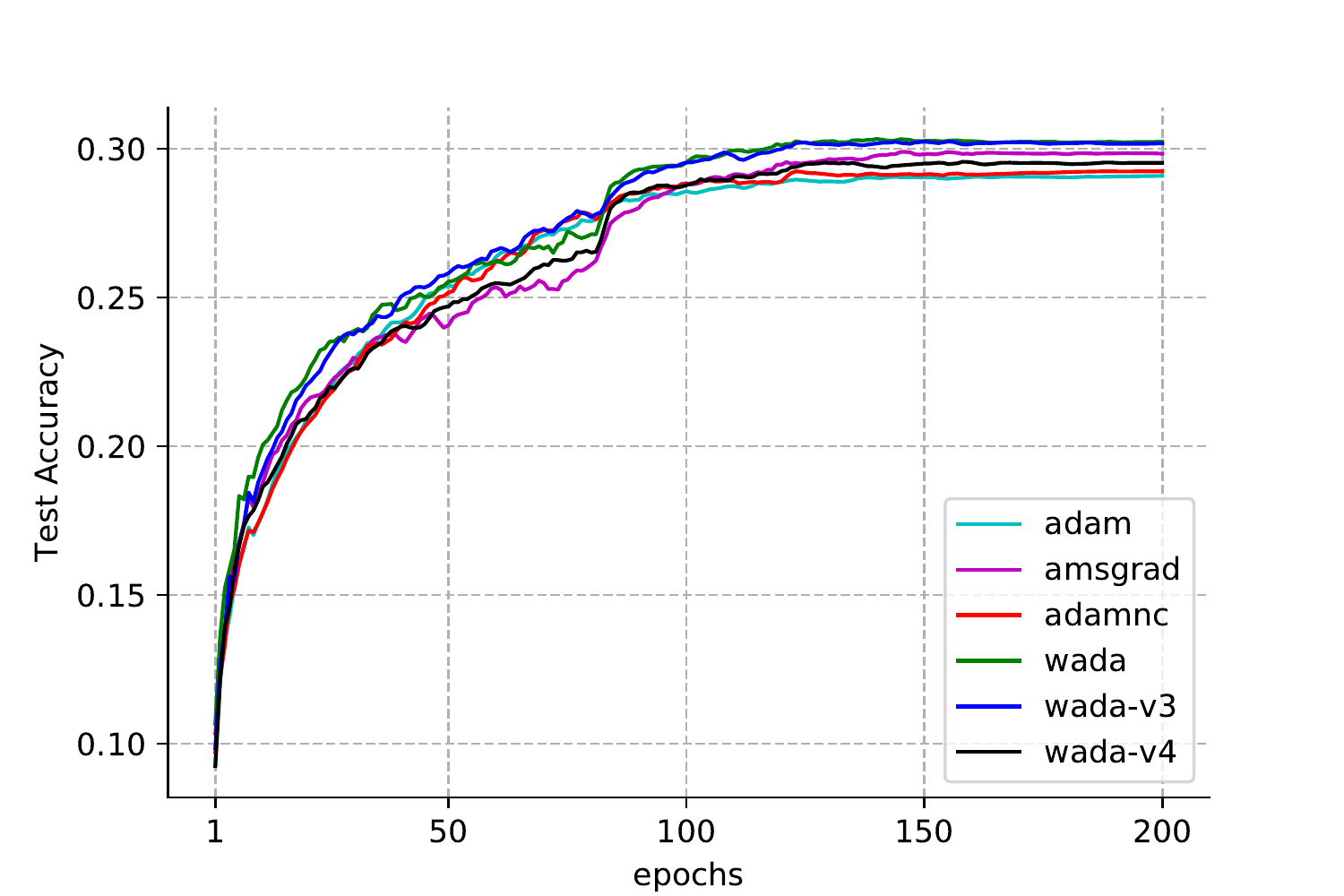}
  }
  \caption{Test Accuracy vs Number of Epoch for 3-layer MLP on MNIST, CIFAR10 and CIFAR100 dataset.}
  \label{fig:acc-mlp}
\end{figure*}

\begin{figure*}[ht] 
  \centering
  \subfigure[CIFAR10 Training Loss]{
    \includegraphics[width=0.4\linewidth]{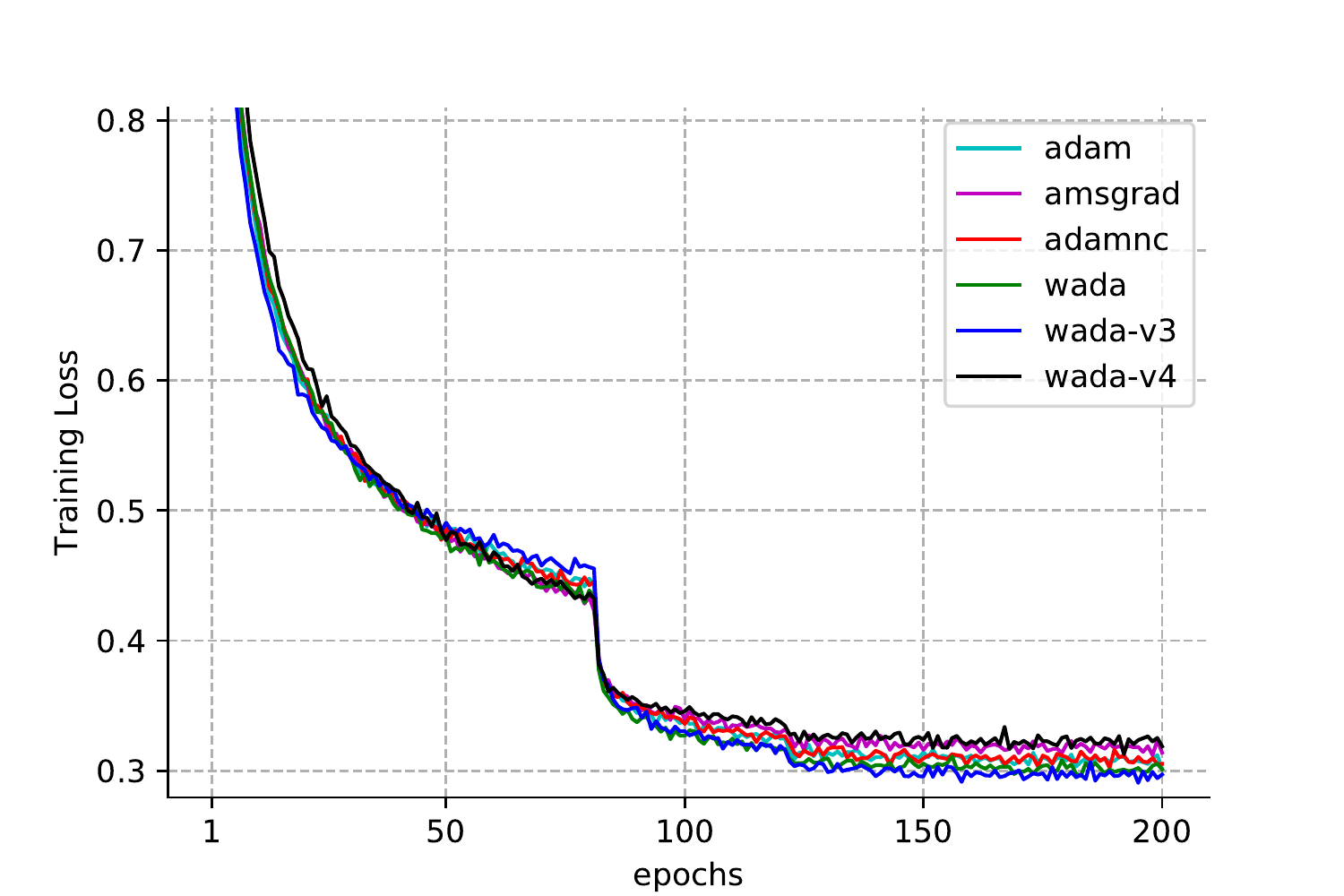}
  }
  \subfigure[CIFAR10 Test Accuracy]{
    \includegraphics[width=0.4\linewidth]{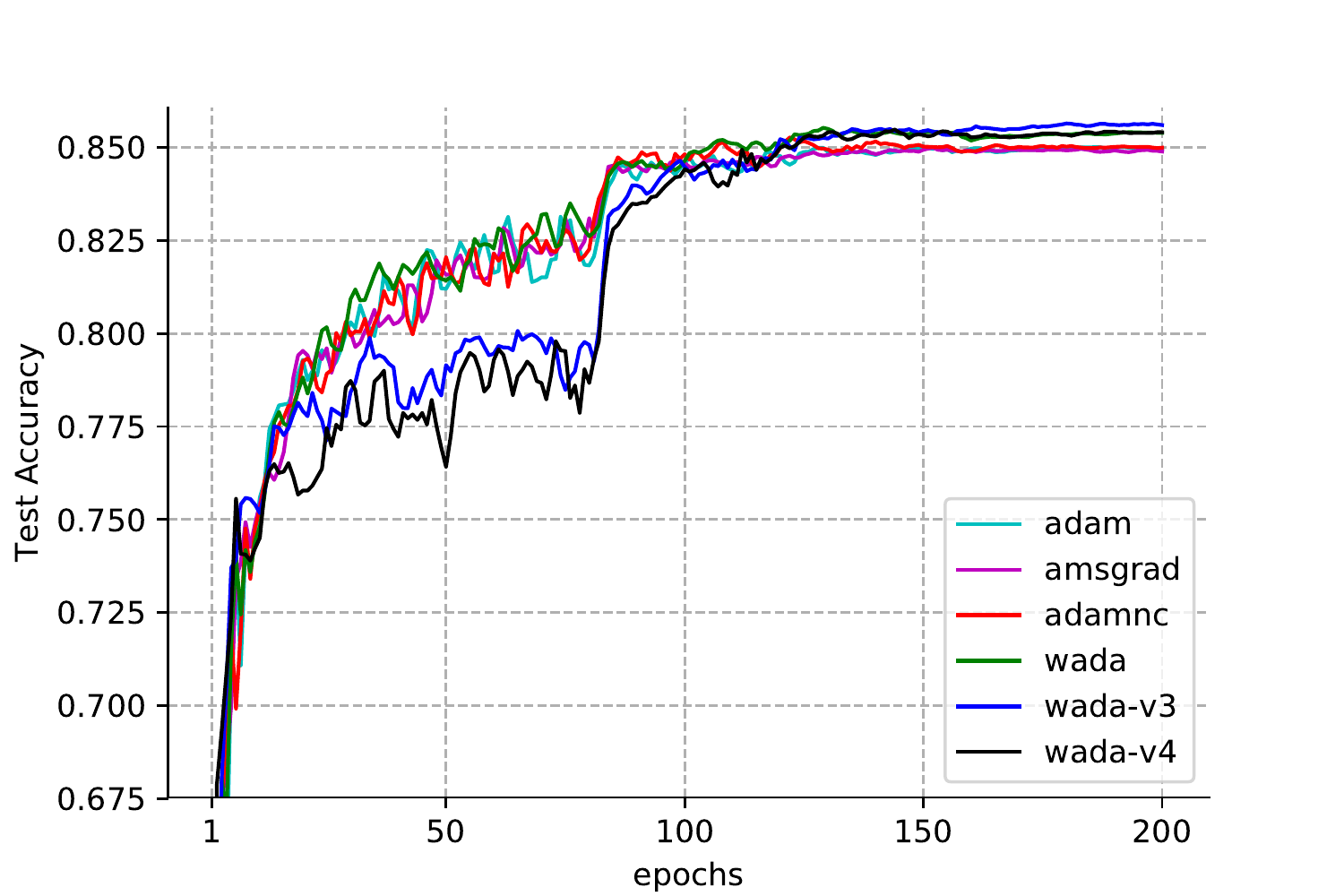}
  }
  \subfigure[CIFAR100 Training Loss]{
    \includegraphics[width=0.4\linewidth]{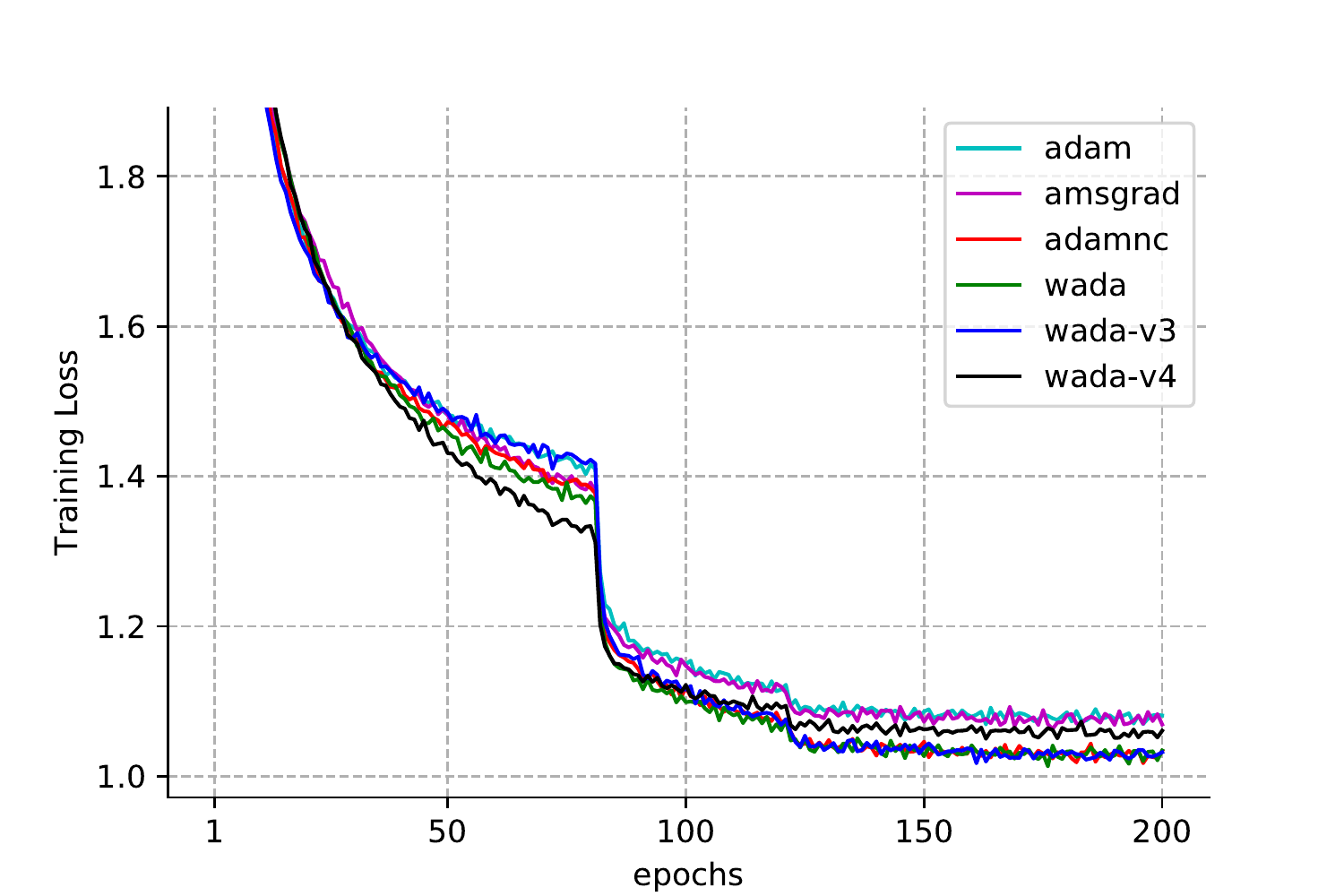}
  }
  \subfigure[CIFAR100 Test Accuracy]{
    \includegraphics[width=0.4\linewidth]{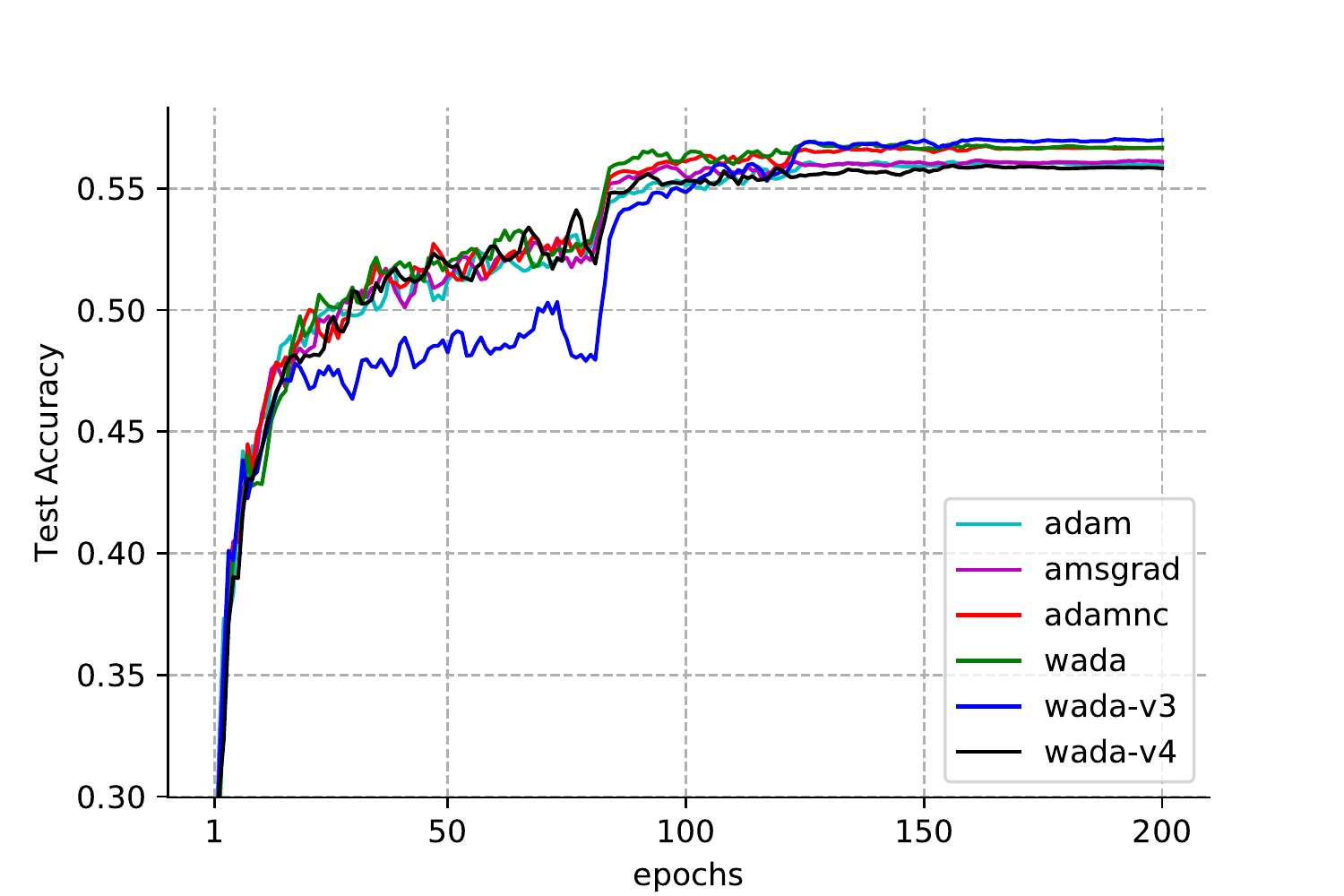}
  }
  \caption{Training Loss and Test Accuracy vs Number of Epoch for Simple CNN on MNIST, CIFAR10 and CIFAR100 dataset.}
  \label{fig:scnn-loss-acc}
\end{figure*}

\begin{figure*}[ht] 
  \centering
  \subfigure[CIFAR10 Training Loss]{
    \includegraphics[width=0.4\linewidth]{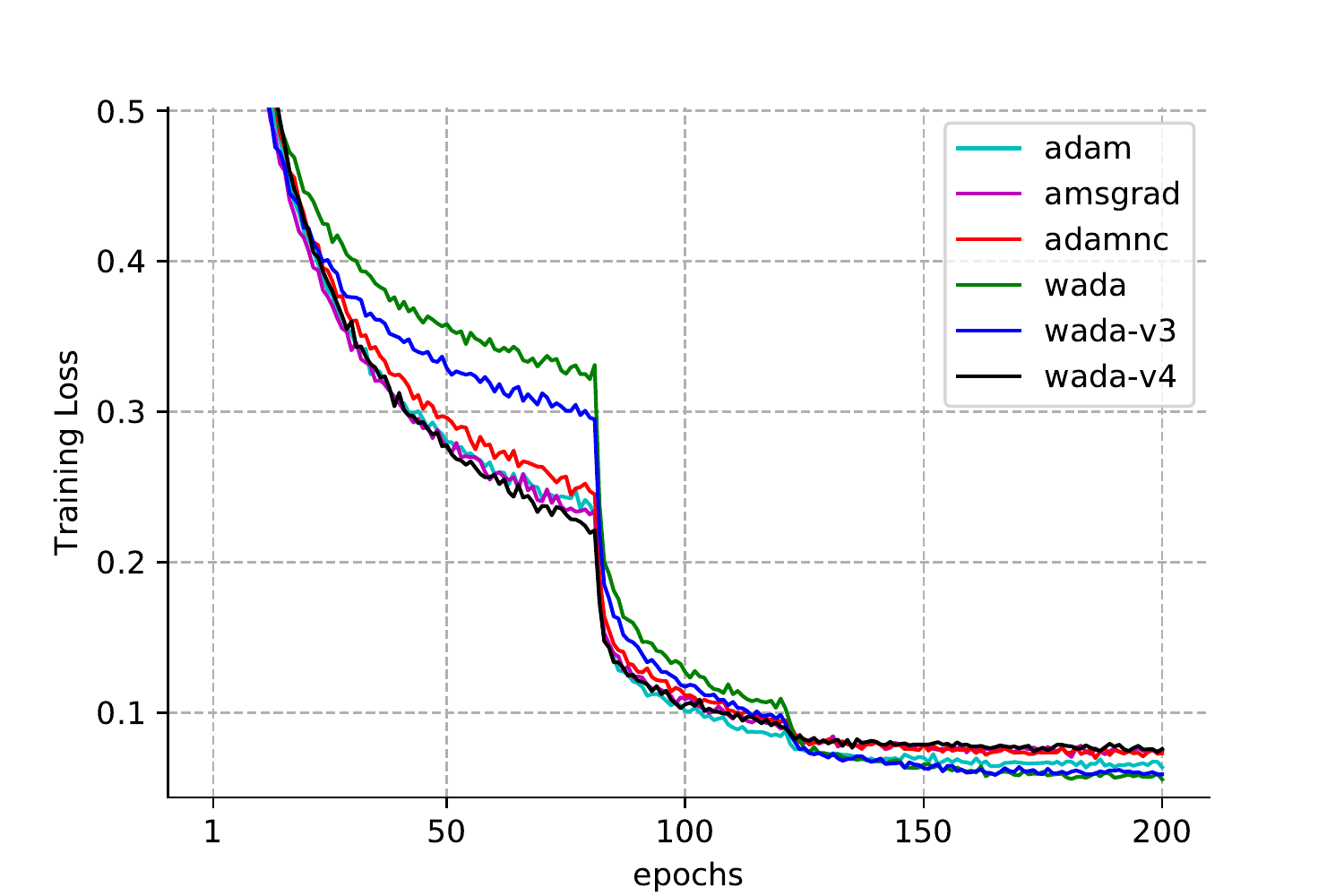}
  }
  \subfigure[CIFAR10 Test Accuracy]{
    \includegraphics[width=0.4\linewidth]{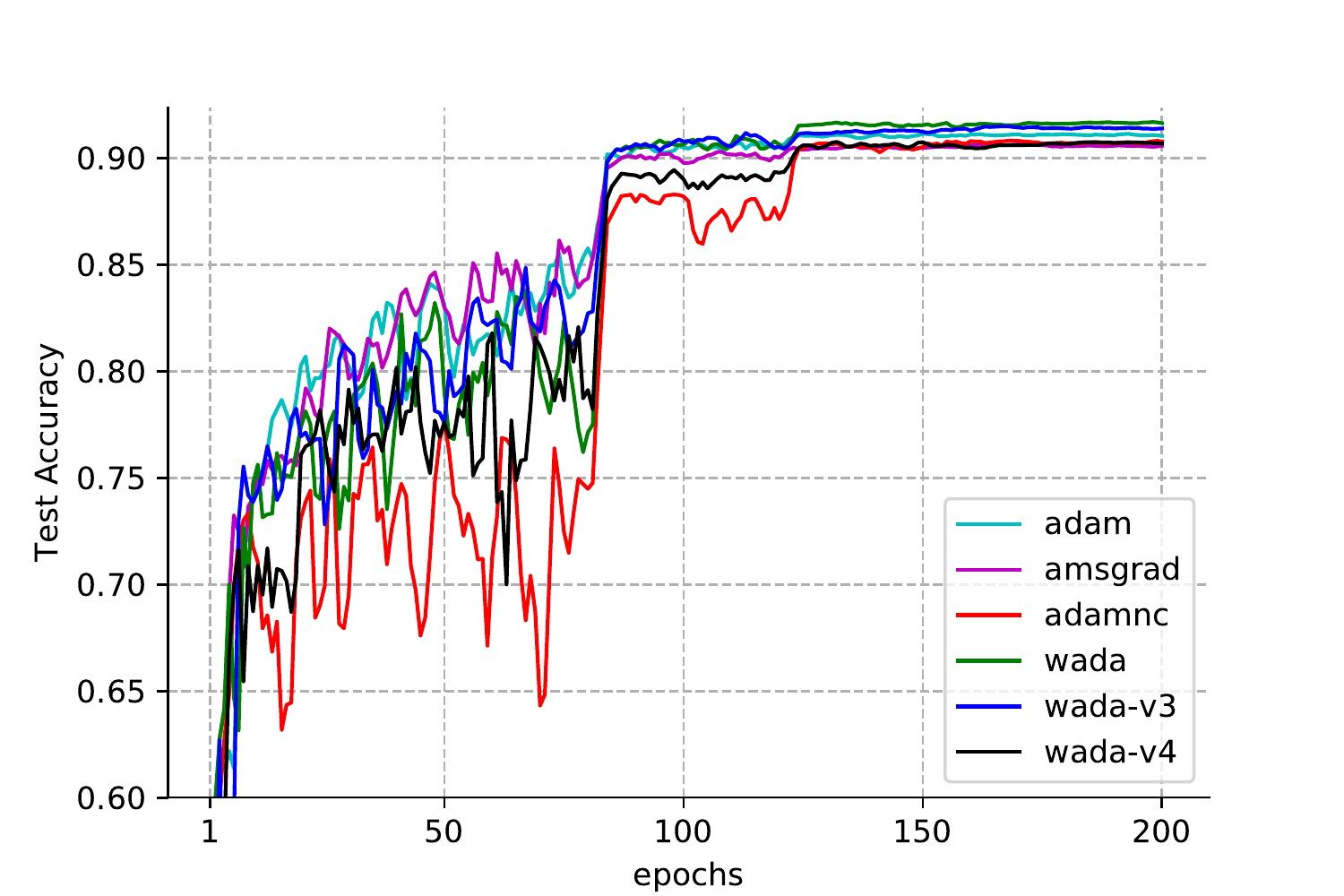}
  }
  \subfigure[CIFAR100 Training Loss]{
    \includegraphics[width=0.4\linewidth]{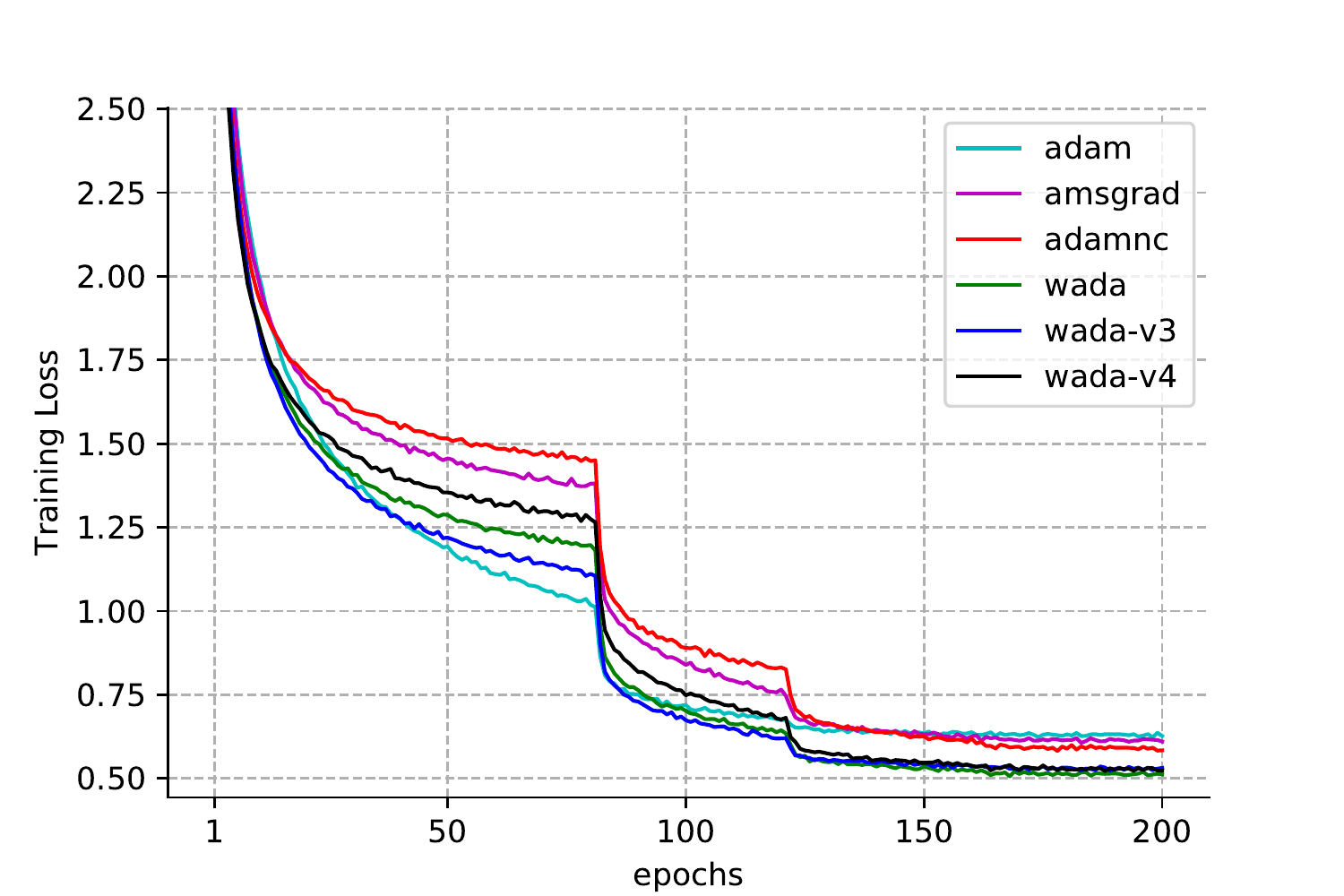}
  }
  \subfigure[CIFAR100 Test Accuracy]{
    \includegraphics[width=0.4\linewidth]{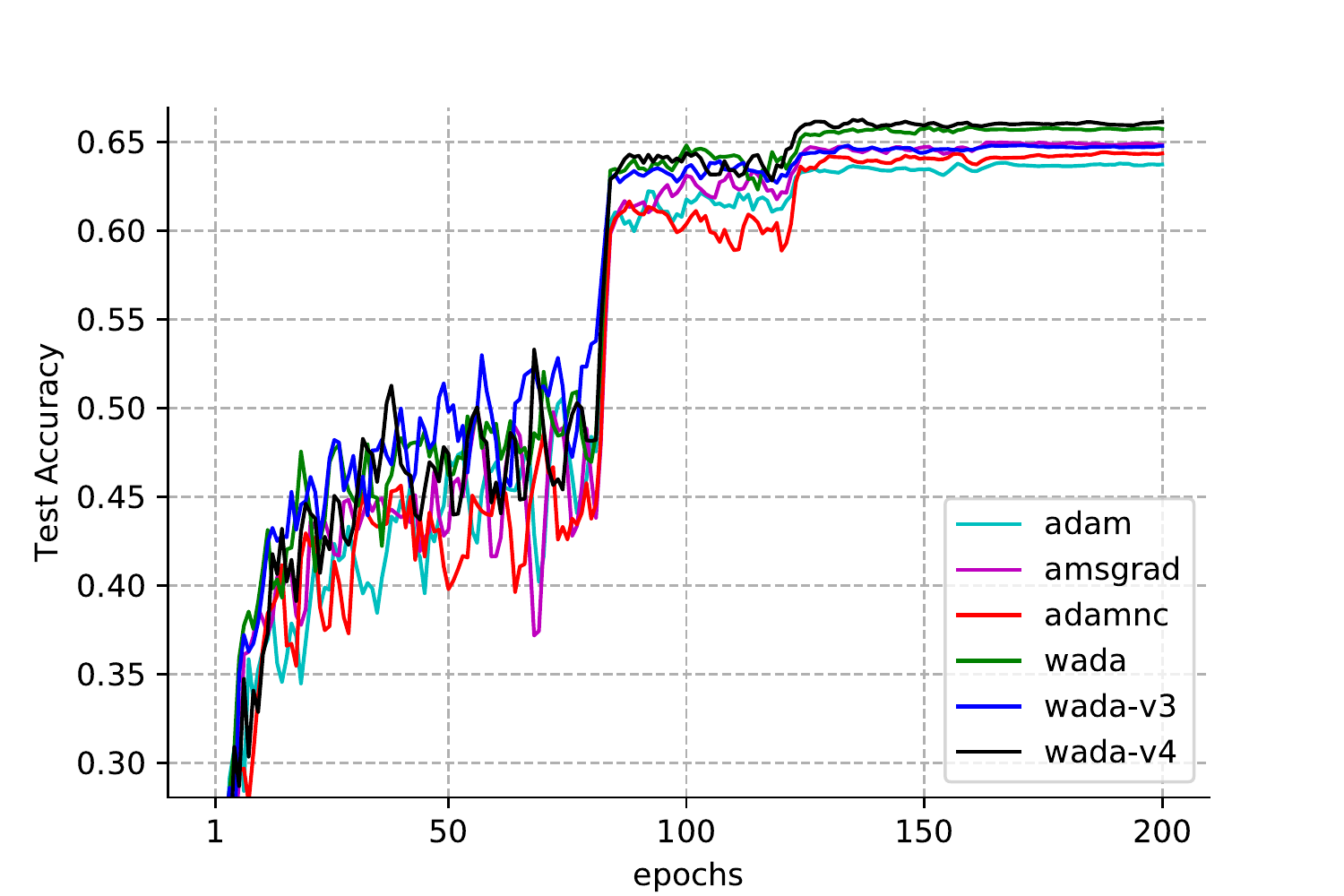}
  }
  \caption{Training Loss and Test Accuracy vs Number of Epoch for ResNet-20 on MNIST, CIFAR10 and CIFAR100 dataset.}
  \label{fig:resnet20-loss-acc}
\end{figure*}

In this section, we mainly evaluate our proposed algorithms in comparison with existing adaptive methods on training convex problems and neural networks. Specifically, we study the multiclass classification problem using softmax regression, multi-layer perceptron and convolutional neural networks. To demonstrate the convergence of our proposed algorithms, we also involve the similar synthetic experiments of the work~\citep{reddi2018convergence}.
All methods are implemented within Keras\footnote{http://keras.io/} (with TensorFlow as the backend) and the experiments are conducted on a Standard Server with Intel Xeon CPU E5-2630 v4 and Nvidia GPU Titan X.

\doublerulesep 0.1pt
\begin{table}[htbp]
  \centering
  \caption{The statistics of the datasets}\label{tab:dataset}
  \begin{footnotesize}
    \begin{tabular*}{\columnwidth}{p{0.23\columnwidth}p{0.23\columnwidth}p{0.23\columnwidth}p{0.23\columnwidth}}
      \toprule 
      Dataset & train samples & test samples & classes \\ 
      \midrule
      MNIST & 60000 & 10000 & 10 \\
      CIFAR10 & 50000 & 10000 & 10 \\
      CIFAR100 & 50000 & 10000 & 100 \\ 
      \bottomrule
    \end{tabular*}
  \end{footnotesize}
\end{table}

\textbf{Datasets:}
We conduct experiments on three popular datasets, i.e., MNIST~\citep{lecun1998gradient}, CIFAR10 and CIFAR100~\citep{krizhevsky2009learning}. The three datasets represent for different difficulty to achieve good performance. Table~\ref{tab:dataset} shows the basic statistical properties of the datasets.

\textbf{Algorithms:}
Since \WADA aims to give a better solution to fix the non-convergence issues of \Adam, we mainly compare with \Adam and its variants in the experiments. 
As \textsc{Padam} and Nostalgic Adam are the variants of \amsgrad or \adamnc, the difference of these algorithms are relatively small. So, we choose the original widely used \amsgrad and  \adamnc as our baselines. Here are some brief introduction of all methods involved in the experiments:
\begin{itemize}
  \item \Adam \citep{kingma2014adam}: The origin \adam have good performance in practice, but may suffer from the non-convergence issues.
  \item \AmsGrad \citep{reddi2018convergence}: \AmsGrad is a variant of \Adam, which maintains the maximum of all past $ v_t $ and uses this maximum value for normalizing the running average of the gradient instead of $ v_t $ in \Adam.
  \item \AdamNc \citep{reddi2018convergence}: \AdamNc is also a variant of \Adam which uses equal weight of squared past gradients in $ v_t $. Actually, it is an momentum base variants of \adagrad.
  \item \WADA (Ours): \WADA is an implementation of \WAGMF which uses the setting of $ \gamma_{t} = t, p_1=2, p_2=4 $.
  \item \threeada (Ours): \threeada is a variant of WADA which uses the setting of $\gamma_{t} = t, p_1=3, p_2=4 $.
  \item \fourada (Ours): \fourada is a variant of \WADA which uses the setting of $\gamma_{t} = t, p_1=4, p_2=4 $.
\end{itemize}

\textbf{Experiment Settings:}
To consistent with the theoretical analysis, we adopt $ O(1/\sqrt{t}) $ step-size decay for all methods on the convex problems. For the nonconvex problem, a constant step-size is used to evaluate the performance in practice. Besides, for better and more stable performance, we use a fixed multi-stage learning rate decaying scheme on nonconvex problem, which is widely used in many works\cite{He:2016tt,Chen:2018tj,yang2018does}. We decay the learning rate by 0.1 at the 80th, 120th and 160th epochs. 
We set $ \beta_{1} = 0.9 $ for all methods and  $ \beta_{2} = 0.999 $ for \Adam and \AmsGrad. These values are most commonly used in practice. The batch size of all experiments is set to 128, and we also add an additional small $ \epsilon = 10^{-7} $ on $ V_t $ to avoid dividing by zero. For better performance and avoid overfitting problem, we use data augmentation \citep{perez2017effectiveness} for all neural network experiments. 
We follow \citep{glorot2010understanding} and randomly initialize parameters with uniform distribution in the range between $ -\sqrt{6/(nin + nout)} $ and $ \sqrt{6/(nin + nout)} $, where $ nin $ and $ nout $ are the numbers of input and output units of the corresponding weight tensor, respectively.
All methods have only one varying parameter: the step-sizes $ \alpha $, which are chose by grid search for all experiments. 

\textbf{Synthetic Experiments:}
To demonstrate the convergence of our algorithm, we conduct the synthetic experiments in the work~\citep{reddi2018convergence}, which use a synthetic example on a simple one-dimensional convex problem. Specifically, we consider the following convex problem:
\begin{equation}
f_t(x) = \left\{
  \begin{array}{lr}
  1010x , & with \ probability \ 0.01 \\
  -10x, & otherwise,
  \end{array}
\right.
\end{equation}
with the constraint set $ \mathcal{F} = [-1,1] $. Obviously, the optimal solution for this problem is $ x = -1 $, and we expect to see whether the algorithms will converge to $ x = -1 $. We demonstrate our results in Figure~\ref{fig:synthetic}. The figures show the average regret ($ R(t)/t $) and the iterate ($ x_t $)  for the problem (for comparison, we also show the $ R(t)/t $ plot for online setting, which sets the first part of the function $ f_t(x)  $  to $ for\ t\ mod\ 101 = 1 $). We can see that the average regret of \adam does not converge to 0 with increasing $ t $, while \AmsGrad and \wada do. Furthermore, for the iterates $ x_t $, \wada and \amsgrad converge very well to the optimal value $ x_t = -1 $. Based on this experiments, we can see that \wada does fix the non-convergence issues of \adam.

\textbf{Softmax Regression: }
\label{sub:lr}
To investigate the performance of our methods on convex problem, we conduct experiments on L2-regularized softmax regression problem. Softmax regression uses a linear model with cross entropy loss and L2-regularized loss. The objective function is defined as
\begin{equation}
F(w) =-\frac{1}{n}\sum_{i=1}^n\text{log}\Bigg(\frac{e^{w_{y_i}^Tx_i + b_{y_i}}}{\sum_{j=1}^Ke^{w_j^Tx_i + b_j}}\Bigg) + \lambda\sum_{k=1}^K \|w_k\|^2 .
\end{equation}
We use the regularization parameter which can achieve the best performance on the test set.
Figure~\ref{fig:loss-lr} shows the training loss results on softmax regression. As shown in the figures, our methods achieve lower training loss. \wada and \fourada perform better than all the other methods in terms of training loss on MNIST dataset. On both CIFAR10 and CIFAR100, our methods outperform all baselines. 

In the following several paragraphs, we mainly focus on the performance of our methods on nonconvex problems. In particular, we conduct experiments on multi-layer perceptron, simple convolutional neural networks and residual neural networks. For the computational complexity of single batch, our methods almost have the same computational time as baselines, i.e., for the following experiments on the CIFAR dataset, approximately 6 ms, 12 ms, 52 ms per batch.

\textbf{Multi-Layer Perceptron:}
We conduct experiments on a simple 3-layer multi-layer perceptron, with 2 hidden full connected layers and 1 softmax layer for the multiclass classification problem on three datasets. The hidden layers have 512 units for each layer in which ReLU activation function \citep{nair2010rectified} and 0.5 dropout \citep{srivastava2014dropout} are used. The number of units in final softmax layer is consistent with the classes of each dataset. The results are shown in Figure~\ref{fig:loss-mlp} and~\ref{fig:acc-mlp}. Since the MNIST dataset is easy to achieve good performance, all methods perform similarly. \wada achieves better performance on CIFAR10 and CIFAR100 for both training loss and test accuracy. Other variants of \wada also achieve good performance.

\textbf{Simple Convolutional Neural Network:}
We also conduct experiments on a simple convolutional neural network. This network is constructed with several layers of convolution, no-linear and pooling units. In particular we first use two convolution layers with 32 channels and kernel size of 3 $ \times $ 3 and followed by 2 $ \times $ 2 max pooling layer. Then we repeat the first part with two 64 channels convolution layers and max pooing layers. Finally, we flatten the output and follow by a softmax layer. The results are shown in Figure~\ref{fig:scnn-loss-acc}. \wada and \threeada achieve better performance on CIFAR10 dataset for both training loss and test accuracy. Though \fourada seems not perform very well on training loss, it achieves better performance on test accuracy. For CIFAR100 dataset, \wada, \adamnc and \threeada achieve better performance.

\textbf{Residual Neural Network:}
Finally, we conduct experiments for Residual Neural Network\citep{He:2016tt} (ResNet). ResNet introduces a novel architecture of convolutional neural networks with residual blocks, which utilize shortcut connections to jump over some layers. 
Such shortcut connections can ease the training of deeper networks.
In this experiments, we use ResNet-20, which contains 2 residual blocks, to train on the CIFAR dataset. We report the results in Figure~\ref{fig:resnet20-loss-acc}. As the results shown in the figures, \wada and \threeada achieve best performance on both training loss and test accuracy. All of our methods achieve better performance on training loss on CIFAR100 dataset. Besides, \wada and \fourada achieve best test accuracy on CIFAR100. Our methods are significantly superior to all baselines on CIFAR10 and CIFAR100 datasets in terms of training loss and test accuracy. 

\textbf{Significance Tests:}
Besides, we also conduct significance tests on our experiments (the detailed results and analyses are shown in the Appendixes A.\ref{sec:significance}). Specifically, we choose the Student's t-test to calculate the p\_values on 8 samples of each experimental result and set the statistical significance threshold to 0.05. As we can see from the Table~\ref{tab:significance}, in most cases, the p\_values < 0.05. 
In terms of these significance tests results and the analyses, we think the performance of our experiments can be considered to be very significant.

Based on these experiments, we can see that \wada not only fixed the non-convergence issues of \adam, but also achieved batter performance on both convex and nonconvex problems, especially deep neural networks. 
Given the performance of  \WADA and its variants, we think they could be powerful competitors among existing adaptive methods and we hope they will be valuable for deep learning.

\section{Conclusion} \label{sec:con}
In this paper, we proposed a general \textit{Weighted Adaptive Gradient Method Framework} (\WAGMF) and a novel \textit{Weighted Adaptive algorithm} (\WADA). Specifically, \WAGMF give a general framework to design new adaptive methods which include many existing algorithms. And \WADA fixed the non-convergence issues of \Adam by applying the linear growing weighting strategy to \WAGMF.
Further, we presented the convergence analysis of \WADA on the Online Convex Optimization problem.  The regret bound of \WADA was in a weighted data-dependent form and can be better than the regret bound of \AdaGrad when the gradients decrease rapidly. This bound may partially explain the good performance of \adam in practice.
Moreover, the experimental results for different models and datasets clearly demonstrated that \WADA and its variants are powerful competitors among existing adaptive methods. 
We hope this work could give another perspective to understand the design of adaptive methods, and suggest good design principles for faster and better stochastic optimization.

\Acknowledgements{We thank the anonymous reviewers for their insightful comments and discussions. This research was partially supported by grants from the National Key Research and Development Program of China
(Grant No. 2018YFB1004300) and the National Natural Science Foundation of China (Grant No. 61703386, 61727809, and U1605251).}

\bibliography{refs}

\begin{thebibliography}{10}

\bibitem{robbins1951stochastic}
Robbins H, Monro S.
\newblock A stochastic approximation method.
\newblock The Annals of Mathematical Statistics, 1951, 22(3): 400--407

\bibitem{duchi2011adaptive}
Duchi J~C, Hazan E, Singer Y.
\newblock Adaptive subgradient methods for online learning and stochastic
  optimization.
\newblock Journal of Machine Learning Research, 2011, 12: 2121--2159

\bibitem{tieleman2012lecture}
Tieleman T, Hinton G.
\newblock Lecture 6.5-rmsprop: Divide the gradient by a running average of its
  recent magnitude.
\newblock COURSERA: Neural networks for machine learning, 2012, 4(2): 26--31

\bibitem{zeiler2012adadelta}
Zeiler M~D.
\newblock Adadelta: an adaptive learning rate method.
\newblock arXiv preprint arXiv:1212.5701, 2012

\bibitem{kingma2014adam}
Kingma D~P, Ba~J.
\newblock Adam: A method for stochastic optimization.
\newblock In: International Conference on Learning Representations.
\newblock 2015

\bibitem{yin2018transcribing}
Yin Y, Huang Z, Chen E, Liu Q, Zhang F, Xie X, Hu~G.
\newblock Transcribing content from structural images with spotlight mechanism.
\newblock In: Proceedings of the 24th ACM SIGKDD International Conference on
  Knowledge Discovery \& Data Mining.
\newblock 2018,  2643--2652

\bibitem{krizhevsky2012imagenet}
Krizhevsky A, Sutskever I, Hinton G~E.
\newblock Imagenet classification with deep convolutional neural networks.
\newblock In: Advances in neural information processing systems.
\newblock 2012,  1097--1105

\bibitem{su2018exercise}
Su~Y, Liu Q, Liu Q, Huang Z, Yin Y, Chen E, Ding C, Wei S, Hu~G.
\newblock Exercise-enhanced sequential modeling for student performance
  prediction.
\newblock In: Thirty-Second AAAI Conference on Artificial Intelligence.
\newblock 2018

\bibitem{liu2018finding}
Liu Q, Huang Z, Huang Z, Liu C, Chen E, Su~Y, Hu~G.
\newblock Finding similar exercises in online education systems.
\newblock In: Proceedings of the 24th ACM SIGKDD International Conference on
  Knowledge Discovery \& Data Mining.
\newblock 2018,  1821--1830

\bibitem{salton1988term}
Salton G, Buckley C.
\newblock Term-weighting approaches in automatic text retrieval.
\newblock Information processing \& management, 1988, 24(5): 513--523

\bibitem{lecun1998gradient}
LeCun Y, Bottou L, Bengio Y, Haffner P.
\newblock Gradient-based learning applied to document recognition.
\newblock Proceedings of the IEEE, 1998, 86(11): 2278--2324

\bibitem{Hazan:2016:IOC:3006427.3006428}
Hazan E.
\newblock Introduction to online convex optimization.
\newblock Found. Trends Optim., 2016, 2(3-4): 157--325

\bibitem{reddi2018convergence}
Reddi S~J, Kale S, Kumar S.
\newblock On the convergence of adam and beyond.
\newblock In: International Conference on Learning Representations.
\newblock 2018

\bibitem{pmlr-v70-mukkamala17a}
Mukkamala M~C, Hein M.
\newblock Variants of {RMSP}rop and {A}dagrad with logarithmic regret bounds.
\newblock In: Proceedings of the 34th International Conference on Machine
  Learning.
\newblock 2017,  2545--2553

\bibitem{rakhlin2012making}
Rakhlin A, Shamir O, Sridharan K.
\newblock Making gradient descent optimal for strongly convex stochastic
  optimization.
\newblock In: Proceedings of the 29th International Conference on Machine
  Learning.
\newblock 2012,  1571--1578

\bibitem{shamir2013stochastic}
Shamir O, Zhang T.
\newblock Stochastic gradient descent for non-smooth optimization: Convergence
  results and optimal averaging schemes.
\newblock In: Proceedings of the 30th International Conference on Machine
  Learning.
\newblock 2013,  71--79

\bibitem{lacoste2012simpler}
Lacoste-Julien S, Schmidt M, Bach F.
\newblock A simpler approach to obtaining an o (1/t) convergence rate for the
  projected stochastic subgradient method.
\newblock arXiv preprint arXiv:1212.2002, 2012

\bibitem{Dean:2012wx}
Dean J, Corrado G, Monga R, Chen K, Devin M, Mao M, Senior A, Tucker P, Yang K,
  Le~Q~V, others .
\newblock Large scale distributed deep networks.
\newblock In: Advances in neural information processing systems.
\newblock 2012,  1223--1231

\bibitem{Chen:2018wy}
Chen Z, Xu~Y, Chen E, Yang T.
\newblock {SADAGRAD:} strongly adaptive stochastic gradient methods.
\newblock In: Proceedings of the 35th International Conference on Machine
  Learning.
\newblock 2018,  912--920

\bibitem{Huang:2018vh}
Huang H, Wang C, Dong B.
\newblock Nostalgic adam: Weighing more of the past gradients when designing
  the adaptive learning rate.
\newblock arXiv preprint arXiv:1805.07557, 2018

\bibitem{Chen:2018tj}
Chen J, Gu~Q.
\newblock Closing the generalization gap of adaptive gradient methods in
  training deep neural networks.
\newblock arXiv preprint arXiv:1806.06763, 2018

\bibitem{Zinkevich:2003tf}
Zinkevich M.
\newblock Online convex programming and generalized infinitesimal gradient
  ascent.
\newblock In: Proceedings of the 20th International Conference on Machine
  Learning.
\newblock 2003,  928--936

\bibitem{cesa2004generalization}
Cesa-Bianchi N, Conconi A, Gentile C.
\newblock On the generalization ability of on-line learning algorithms.
\newblock IEEE Transactions on Information Theory, 2004, 50(9): 2050--2057

\bibitem{Bernstein:2018wn}
Bernstein J, Wang Y, Azizzadenesheli K, Anandkumar A.
\newblock {SIGNSGD:} compressed optimisation for non-convex problems.
\newblock In: Proceedings of the 35th International Conference on Machine
  Learning.
\newblock 2018,  559--568

\bibitem{dozat2016incorporating}
Dozat T.
\newblock Incorporating nesterov momentum into adam.
\newblock In: International Conference on Learning Representations, Workshop
  Track.
\newblock 2016

\bibitem{krizhevsky2009learning}
Krizhevsky A, Hinton G.
\newblock Learning multiple layers of features from tiny images.
\newblock Technical report, Citeseer, 2009

\bibitem{He:2016tt}
He~K, Zhang X, Ren S, Sun J.
\newblock Deep residual learning for image recognition.
\newblock In: Proceedings of the IEEE conference on computer vision and pattern
  recognition.
\newblock 2016,  770--778

\bibitem{yang2018does}
Yang T, Yan Y, Yuan Z, Jin R.
\newblock Why does stagewise training accelerate convergence of testing error
  over sgd?
\newblock arXiv preprint arXiv:1812.03934, 2018

\bibitem{perez2017effectiveness}
Perez L, Wang J.
\newblock The effectiveness of data augmentation in image classification using
  deep learning.
\newblock arXiv preprint arXiv:1712.04621, 2017

\bibitem{glorot2010understanding}
Glorot X, Bengio Y.
\newblock Understanding the difficulty of training deep feedforward neural
  networks.
\newblock In: Proceedings of the thirteenth international conference on
  artificial intelligence and statistics.
\newblock 2010,  249--256

\bibitem{nair2010rectified}
Nair V, Hinton G~E.
\newblock Rectified linear units improve restricted boltzmann machines.
\newblock In: Proceedings of the 27th International Conference on Machine
  Learning.
\newblock 2010,  807--814

\bibitem{srivastava2014dropout}
Srivastava N, Hinton G~E, Krizhevsky A, Sutskever I, Salakhutdinov R.
\newblock Dropout: a simple way to prevent neural networks from overfitting.
\newblock Journal of Machine Learning Research, 2014, 15(1): 1929--1958

\end{thebibliography}
\bibliographystyle{fcs}
\appendix
\section{Appendixes}

\subsection{\textbf{Significance test results analysis}}\label{sec:significance}
\begin{table*}[htbp]
	\caption[]{\begin{minipage}[t]{.93\linewidth}Significance test results (p\_values) for Figure.\ref{fig:loss-lr} to  Figure.\ref{fig:resnet20-loss-acc}.  All significance tests are conduct on Student's t-test, and the statistical significance threshold is set to 0.05. In the table, the p\_values below the significance threshold are shown in black font, otherwise gray.\end{minipage}}
	\label{tab:significance}
	\begin{footnotesize}
		\begin{tabular}{p{50pt}|p{37pt}p{37pt}p{37pt}p{37pt}p{37pt}p{37pt}p{37pt}p{37pt}p{37pt}}
			\toprule
			Baseline vs. &  \adam  & \adam  & \adam  &  \amsgrad  & \amsgrad   & \amsgrad&  \adamnc\ & \adamnc & \adamnc \\ 
			Ours &  \wada & \threeada  & \fourada & \wada &  \threeada  &  \fourada & \wada &  \threeada  &  \fourada\\ 
			\midrule
			Figure \ref{fig:loss-lr}.~(a) & 	6.79e-08 & 	\textcolor[rgb]{0.5,0.5,0.5}{3.81e-01} & 	5.28e-07 & 	1.55e-07 & 	\textcolor[rgb]{0.5,0.5,0.5}{1.37e-01} & 	4.46e-06 & 7.02e-11 & 	\textcolor[rgb]{0.5,0.5,0.5}{2.93e-01} & 	6.97e-09\\
			Figure \ref{fig:loss-lr}.~(b)	 & 	5.61e-09 & 	1.11e-13 & 	1.17e-08 & 	\textcolor[rgb]{0.5,0.5,0.5}{9.56e-02} & 	2.05e-12 & 	7.01e-05 & 	4.51e-03 & 	1.66e-10 & 	2.30e-05\\
			Figure \ref{fig:loss-lr}.~(c)	 & 	1.49e-12 & 	2.68e-11 & 	1.19e-11 & 	1.23e-09 & 	1.00e-07 & 	2.60e-08 & 	2.32e-06 & 	2.86e-04 & 	5.26e-05\\
			Figure \ref{fig:loss-mlp}.~(a)	 & 	\textcolor[rgb]{0.5,0.5,0.5}{7.72e-01} & 	1.21e-02 & 	\textcolor[rgb]{0.5,0.5,0.5}{1.07e-01} & 	\textcolor[rgb]{0.5,0.5,0.5}{7.83e-01} & 	1.21e-02 & 	1.04e-02 & 	\textcolor[rgb]{0.5,0.5,0.5}{8.24e-01} & 	1.47e-02 & 	1.43e-02\\
			Figure \ref{fig:loss-mlp}.~(b)	 & 	3.61e-07 & 	\textcolor[rgb]{0.5,0.5,0.5}{2.72e-01} & 	\textcolor[rgb]{0.5,0.5,0.5}{8.19e-02} & 	\textcolor[rgb]{0.5,0.5,0.5}{1.85e-01} & 	4.01e-04 & 	1.56e-03 & 	3.89e-06 & 	1.58e-01 & 	\textcolor[rgb]{0.5,0.5,0.5}{5.86e-02}\\
			Figure \ref{fig:loss-mlp}.~(c)	 & 	3.02e-12 & 	8.41e-06 & 	\textcolor[rgb]{0.5,0.5,0.5}{2.08e-01} & 	2.64e-09 & 	2.14e-04 & 	3.63e-10 & 	9.69e-15 & 	3.42e-10 & 	1.92e-08\\
			Figure \ref{fig:acc-mlp}.~(a)	 & 	6.61e-11 & 	4.01e-02 & 	1.37e-09 & 	0.00e+00 & 	2.34e-06 & 	0.00e+00 & 	1.25e-19 & 	3.73e-14 & 	2.34e-06\\
			Figure \ref{fig:acc-mlp}.~(b)	 & 	9.57e-24 & 	2.78e-18 & 	5.64e-16 & 	1.34e-18 & 	4.01e-14 & 	2.65e-17 & 	5.13e-20 & 	8.16e-14 & 	2.71e-10\\
			Figure \ref{fig:acc-mlp}.~(c)	 & 	8.09e-29 & 	2.40e-26 & 	1.77e-21 & 	3.26e-19 & 	2.99e-17 & 	3.09e-17 & 	3.06e-28 & 	1.48e-25 & 	4.55e-19\\
			Figure \ref{fig:scnn-loss-acc}.~(a)	 & 	8.60e-04 & 	8.41e-06 & 	1.85e-08 & 	4.65e-08 & 	7.34e-09 & 	1.86e-03 & 	8.64e-05 & 	1.24e-06 & 	1.45e-09\\
			Figure \ref{fig:scnn-loss-acc}.~(b)	 & 	1.85e-15 & 	1.97e-16 & 	3.58e-15 & 	1.51e-16 & 	3.70e-17 & 	2.91e-16 & 	4.68e-15 & 	3.56e-16 & 	8.96e-15\\
			Figure \ref{fig:scnn-loss-acc}.~(c)	 & 	3.72e-10 & 	1.09e-11 & 	5.54e-07 & 	1.63e-09 & 	9.11e-11 & 	7.57e-06 & 	\textcolor[rgb]{0.5,0.5,0.5}{9.65e-01} & 	\textcolor[rgb]{0.5,0.5,0.5}{6.46e-01} & 	1.93e-08\\
			Figure \ref{fig:scnn-loss-acc}.~(d)	 & 	2.11e-19 & 	1.48e-21 & 	1.78e-09 & 	5.46e-15 & 	1.14e-17 & 	7.34e-11 & 	\textcolor[rgb]{0.5,0.5,0.5}{3.42e-01} & 	2.31e-14 & 	4.70e-20\\
			Figure \ref{fig:resnet20-loss-acc}.~(a)	 & 	3.68e-09 & 	2.36e-08 & 	2.71e-10 & 	6.18e-15 & 	9.44e-16 & 	\textcolor[rgb]{0.5,0.5,0.5}{9.69e-02} & 	7.13e-13 & 	7.10e-13 & 	2.56e-03\\
			Figure \ref{fig:resnet20-loss-acc}.~(b)	 & 	4.49e-13 & 	6.79e-11 & 	4.22e-12 & 	3.30e-17 & 	4.50e-17 & 	2.47e-07 & 	4.27e-14 & 	8.77e-13 & 	\textcolor[rgb]{0.5,0.5,0.5}{1.37e-01}\\
			Figure \ref{fig:resnet20-loss-acc}.~(c)	 & 	1.81e-18 & 	2.21e-17 & 	3.01e-17 & 	5.88e-19 & 	1.23e-17 & 	2.18e-17 & 	3.73e-16 & 	1.71e-14 & 	1.90e-14\\
			Figure \ref{fig:resnet20-loss-acc}.~(d)	 & 	1.29e-19 & 	2.41e-15 & 	2.48e-18 & 	9.62e-16 & 	1.88e-05 & 	9.93e-15 & 	6.07e-19 & 	3.33e-11 & 	3.98e-17\\
			\bottomrule 
		\end{tabular}
	\end{footnotesize}
\end{table*}
We use the Student's t-test to carry out significance tests on the experimental results. Specifically,
we choose 8 samples of each experiment's result to calculate the p\_value, and set the statistical significance threshold to 0.05.
The p\_values of the significance tests are shown in Table~\ref{tab:significance}.
As we can see from the Figure~\ref{fig:loss-lr}. (a) and the Table~\ref{tab:significance}, though \threeada performs the same with baselines, \wada and \fourada significantly perform better than all baselines. For the performance in Figure~\ref{fig:loss-mlp}. (a), because of the MNIST dataset is easy to achieve good performance, our methods and baselines both perform well. Therefore, the p\_values of significance tests are relatively large.
And for the simple CNN and ResNet-20 experiments' results in Figure~\ref{fig:scnn-loss-acc} and  Figure~\ref{fig:resnet20-loss-acc}, our methods significantly  outperform the baselines, and most of the p\_values on the experiments are very significant.

As we can see from the results and analyses, in most cases, the p\_values < 0.05. 
Based on these significance tests, 
we think that the performance of our experiments can be considered to be significant.

\subsection{\textbf{Lemmas}}

\begin{lemma}\label{lem:convex}
  Let function $ f: \R^d \to \R $ be convex, then for all $ x,y \in \R^d , g(x) \in \partial f(x) $, then
  \[
  f(y) \ge f(x) + g(x)^T (y-x) .
  \]
\end{lemma}

\begin{lemma}\label{lem:projection}
  (Refer from \citep{pmlr-v70-mukkamala17a}) Let $ V \in \mathcal{S}_d^{+}  $ be a symmetric, positive definite matrix and $ \mathcal{F} \in \R^d $ be a convex set, then
  \[
  \norm{ P_{\mathcal{F}}^{V} (x) -   P_{\mathcal{F}}^{V} (y)  }_V \le \norm{x - y}_V .
  \]
\end{lemma}

\begin{lemma} \label{lem:sum-r}
  Let $ M \in \R ,x_i \in \R $, $ 1 \le M $ and  $ 0 \le {x_i} \le M^2 $, we have
  \begin{equation*}
  \sum_{i=1}^{n}  \frac{
    x_i
  }{
    \sqrt[4]{\sum_{j=1}^{i} j  \cdot  x_j}  
  } \le M \cdot  \sqrt[4]{\sum_{i=1}^{n} i  \cdot  x_i} .
  \end{equation*}
  
  \begin{proof}
    The lemma is clearly true for $ n=1 $. Fix some $ n $, and we assume the lemma holds for $ n-1 $, that is
    \begin{equation*}
    \sum_{i=1}^{n-1}  \frac{
      x_i
    }{
      \sqrt[4]{\sum_{j=1}^{i} j  \cdot  x_j}  
    } \le M \cdot  \sqrt[4]{\sum_{i=1}^{n-1} i  \cdot  x_i}.
    \end{equation*}
    Thus, we define $ Z = \sqrt{ \sum_{i=1}^{n} i  \cdot  x_i } $ and $ x = x_n $, we have
    \begin{align*}
    \ & \sum_{i=1}^{n}  \frac{
      x_i
    }{
      \sqrt[4]{\sum_{j=1}^{i} j  \cdot  x_j}  
    } 
    \\& \le M \cdot  \sqrt[4]{\sum_{i=1}^{n-1} i  \cdot  x_i}
    + \frac{x_n}{
      \sqrt[4]{\sum_{i=1}^{n} i  \cdot  x_i}
    }
    \\& \le M \cdot  \sqrt[4]{
      Z^2 - n x
    }
    + \frac{x}{
      \sqrt{Z}
    } .
    \end{align*}
    The derivative of the right hand side with respect to $ x $ is
    \begin{align*}
    D(x_n) =  -nM \cdot (Z^2-nx)^{-3/4}    + Z^{-1/2}.
    \end{align*}
    For $ x = 0 $, $ D(0) = Z^{-1/2} (1- Mn/Z ) $, for $ Z \le \sqrt{n^2 M^2} $, we have $ D(0) \le 0 $. For $ x > 0 $, $ D(x) < D(0) < 0 $. So, the derivative is negative for $ x>0 $. Thus, subject to the constraint $ x \ge 0 $, the right hand side is maximized at $ x = 0 $, and is therefore at most $  M \cdot \sqrt[4]{ \sum_{i=1}^{n} i  \cdot  x_i}$.
  \end{proof}
\end{lemma}

\subsection{\textbf{Proof of Theorem~\ref{thm:conv}}}
The proof of the Theorem \ref{thm:conv}  follows the convergence analysis of \AdaGrad \citep{duchi2011adaptive} and Theorem 4 in \citep{reddi2018convergence}. 
\begin{proof}
  We assume that $ x^* \in \R^d $ is the optimal point of the problem. 
  Follow the update rules in algorithm~\ref{alg:wagmf}, we have
  \begin{align*}
  &\norm{x_{t+1} - x^*}_{V_t}^2 
  = \norm{P_{\mathcal{F}}^{V_t} (x_t - \alpha_t V_t^{-1}  m_t )  - x^*}_{V_t}^2 
  \\&\le \norm{x_t - \alpha_t V_t^{-1}  m_t   -   x^*}_{V_t}^2
  \\& = \norm{x_t - x^*}_{V_t}^2  + \alpha_t^2 \langle m_t , V_t^{-1} m_t \rangle - 2\alpha_t  \langle m_t , x_t - x^* \rangle
  \\& =  \norm{x_t - x^*}_{V_t}^2 
    + \alpha_t^2 \langle m_t , V_t^{-1} m_t \rangle 
    \\&- 2\alpha_t \langle \beta_{1t} m_{t-1} + (1-\beta_{1t}) g_t  , x_t - x^* \rangle.
  \end{align*}
  Since $P_{\mathcal{F}}^{V_t} (x^*)  = x^* $ ,we can get the first inequality and by applying lemma \ref{lem:projection}. Then, by rearrange the above inequality we can get
  \begin{equation}\label{equ:grad}
  \begin{aligned}
  <&g_t,x_t-x_*> 
  \\& \le \frac{1}{2 \alpha_t(1-\beta_{1t})} \mbra{
    \norm{x_{t} - x^*}_{V_t}^2  -  \norm{x_{t+1} - x^*}_{V_t}^2 
  }  \\&+ \frac{\alpha_t}{2 (1-\beta_{1t})} \langle m_t , V_t^{-1} m_t \rangle
  + \frac{\beta_{1t} }{1-\beta_{1t}} \langle m_{t-1}, x_{t-1} - x^* \rangle
  \\& \le \frac{1}{2 \alpha_t(1-\beta_{1t})} \mbra{
    \norm{x_{t} - x^*}_{V_t}^2  -  \norm{x_{t+1} - x^*}_{V_t}^2 
  }  
  \\& + \frac{\alpha_t}{2 (1-\beta_{1t})}  \norm{ m_t}_{V_t^{-1}}^2
  + \frac{\beta_{1t} \alpha_t }{2 (1-\beta_{1t})} \norm{m_{t-1}}_{V_{t-1}^{-1}}^2 
  \\& + \frac{\beta_{1t}  }{2\alpha_t (1-\beta_{1t})} \norm{x_t - x_*}_{V_{t-1}}^2  .
  \end{aligned} 
  \end{equation}
  The second inequality follows Cauchy-Schwarz inequality. Hence we can upper bound the regret by applying above inequality:
  \begin{equation}\label{equ:main}
  \begin{aligned}
  & R(T)  = \sum_{t=1}^{T}  f_t(x_t) -f_t(x^*) \le \sum_{t=1}^{T} <g_t,x_t-x_*> 
  \\& \le \sum_{t=1}^{T}
  \left[  \frac{1}{2 \alpha_t(1-\beta_{1t})} \mbra{
    \norm{x_{t} - x^*}_{V_t}^2  -  \norm{x_{t+1} - x^*}_{V_t}^2  }  
  \right. \\& \left. 
  + \frac{\alpha_t}{2 (1-\beta_{1t})} \norm{ m_t}_{V_t^{-1}}^2  
  + \frac{\beta_{1t} \alpha_t }{2 (1-\beta_{1t})} \norm{m_{t-1}}_{V_{t-1}^{-1}}^2
  \right. \\& \left. 
  + \frac{\beta_{1t}  }{2\alpha_t (1-\beta_{1t})} \norm{x_t - x_*}_{V_{t-1}}^2  \right] .
  \end{aligned}
  \end{equation}
  By applying $ \beta_{1t} \le \beta_{1} \le 1 $, we have
  \begin{align*}
  & R(T)  
  \le \sum_{t=1}^{T}
    \left[  \frac{1}{2 \alpha_t(1-\beta_{1t})} \mbra{
      \norm{x_{t} - x^*}_{V_t}^2  -  \norm{x_{t+1} - x^*}_{V_t}^2  }
    \right. \\& \left.  
    + \frac{\beta_{1t}  }{2\alpha_t (1-\beta_{1t})} \norm{x_t - x_*}_{V_{t-1}}^2  \right] 
     + \sum_{t=1}^{T}  \frac{\alpha_t}{1-\beta_1} \norm{m_t}_{V_t^{-1}}^2
  \\& \le \frac{1}{2 \alpha_1(1-\beta_{1})} \norm{x_{1} - x^*}_{V_1}^2 
    \\& + \sum_{t=2}^{T} \frac{1}{2 (1-\beta_{1})} \mbra{
      \frac{\norm{x_{t} - x^*}_{V_t}^2}{\alpha_t}  - \frac{ \norm{x_{t} - x^*}_{V_{t-1}}^2 }{\alpha_{t-1}} }  
    \\& +  \sum_{t=1}^{T} \frac{\beta_{1t}}{2\alpha_t (1-\beta_{1t})} 
    \norm{x_t - x_*}_{V_{t-1}}^2 
    + \sum_{t=1}^{T}  \frac{\alpha_t}{1-\beta_1} \norm{m_t}_{V_t^{-1}}^2 .
  \end{align*}
  So, since $ V_{t} =\sqrt[p_2]{v_t  \cdot  b_t} $, $ v_t \ge v_{t-1}$ and $ b_t^{-p_2}/\alpha_{t} \ge  b_{t-1}^{-p_2}/ \alpha_{t-1} $, we have $ V_{t,i}/\alpha_{t} \ge V_{t-1,i}/\alpha_{t-1} $. Then
  \begin{align*}
  &R(T)  
  \le 
  \frac{1}{2 \alpha_1(1-\beta_{1})} 
  \sum_{i=1}^{d}  V_1 (x_{1,i} - x_i^*)^2  
  \\& +  \frac{1}{2  (1-\beta_{1})} \sum_{t=2}^{T} \sum_{i=1}^{d}
  (x_{t,i} - x_i^*)^2 
  \mbra{
    \frac{ V_{t,i}}{\alpha_{t}} - \frac{V_{t-1,i}}{\alpha_{t-1}}
  }
  \\& + \sum_{t=1}^{T} \sum_{i=1}^{d} \frac{
    \beta_{1t} (x_{t,i} - x_i^*)^2 V_{t-1,i}
  }{
    2\alpha_t (1-\beta_{1t})
  } 
    + \sum_{t=1}^{T} \frac{\alpha_t}{1-\beta_1} \norm{m_t}_{V_t^{-1}}^2
  \\
  &\le 
  \frac{D_\infty^2}{2 \alpha_T (1-\beta_{1})} \sum_{i=1}^{d}  V_{T,i} 
  + \frac{D_\infty^2}{2 } \sum_{t=1}^{T} \sum_{i=1}^{d}
   \frac{\beta_{1t}  V_{t-1,i} }{ (1-\beta_{1t}) \alpha_t }
  \\& + \sum_{t=1}^{T}  \frac{\alpha_t}{1-\beta_1} \norm{m_t}_{V_t^{-1}}^2 .
  \end{align*}
  The last inequality is using the assumption $  \norm{x_T - x^*}_\infty \le D_\infty $. 
\end{proof}

\subsection{\textbf{Proof of Theorem~\ref{thm:equ-w}}}
\begin{proof}
  Based on the result of Theorem~\ref{thm:conv}, we further bound the three terms in Theorem~\ref{thm:conv}.
  \begin{align*}
  R(T) &\le 
  \frac{D_\infty^2}{2 \alpha_T (1-\beta_{1})} \sum_{i=1}^{d}  V_{T,i} 
  + \frac{D_\infty^2}{2 } \sum_{t=1}^{T} \sum_{i=1}^{d}
  \frac{\beta_{1t}  V_{t-1,i} }{ (1-\beta_{1t}) \alpha_t }
  \\& + \sum_{t=1}^{T}  \frac{\alpha_t}{1-\beta_1} \norm{m_t}_{V_t^{-1}}^2 .
  \end{align*}
  First, we bound the last term $ \sum_{t=1}^{T}  \frac{\alpha_t}{1-\beta_1} \norm{m_t}_{V_t^{-1}}^2$. let $ \alpha_t = \frac{\alpha}{\sqrt{t}} $ ,$ V_{t,i} = t^{-1/p} \sqrt[p]{ \sum_{j=1}^{t} g_{j,i}^p } $, we have
  \begin{align*}
   &\sum_{t=1}^{T}  \frac{\alpha_t}{1-\beta_1} \norm{m_t}_{V_t^{-1}}^2
   \\& = \sum_{t=1}^{T-1}  \frac{\alpha_t}{1-\beta_1} \norm{m_t}_{V_t^{-1}}^2 +  \frac{\alpha}{1-\beta_1} \sum_{i=1}^{d} \frac{m_{T,i}}{ \sqrt[p]{ T^{p/2-1}  v_{T,i}}}   .
  \end{align*}
  We can further bound the $ m_{T,i}^2 $ term by applying Cauchy-Schwarz inequality:
  \begin{equation}
  \begin{aligned}
    m_{T,i}^2 &= (\sum_{j=1}^{T} \prod_{k=1}^{T-j} \beta_{1(T-k+1)} g_{j,i} )^2 
    \\& \le ( \sum_{j=1}^{T} \prod_{k=1}^{T-j} \beta_{1(T-k+1)} )(\sum_{j=1}^{T} \prod_{k=1}^{T-j} \beta_{1(T-k+1)} g_{j,i}^2)
    \\& \le (\sum_{j=1}^{T}  \beta_{1}^{T-j}) (\sum_{j=1}^{T}  \beta_{1}^{T-j} g_{j,i}^2 )
    \le  \frac{1}{ 1-\beta_{1} }  \sum_{j=1}^{T}  \beta_{1}^{T-j} g_{j,i}^2 .
  \end{aligned}
  \end{equation}
  The second inequality is for $ \beta_{1t} \le \beta_{1} $. Now, we have:
  \begin{align*}
  &\sum_{t=1}^{T}  \frac{\alpha_t}{1-\beta_1} \norm{m_t}_{V_t^{-1}}^2
  \\& \le \sum_{t=1}^{T-1}  \frac{\alpha_t}{1-\beta_1} \norm{m_t}_{V_t^{-1}}^2 +  \frac{\alpha}{(1-\beta_1)^2} \sum_{i=1}^{d} \frac{ \sum_{j=1}^{T}  \beta_{1}^{T-j} g_{j,i}^2 }{ \sqrt[p]{ T^{p/2-1}  v_{T,i}}}  
  \\& \le \sum_{t=1}^{T-1}  \frac{\alpha_t}{1-\beta_1} \norm{m_t}_{V_t^{-1}}^2 +  \frac{\alpha}{(1-\beta_1)^2} \sum_{i=1}^{d}  \sum_{j=1}^{T}  \frac{ \beta_{1}^{T-j} g_{j,i}^2 }{ \sqrt[p]{ j^{p/2-1}  v_{j,i}}}
  \\& \le  \frac{\alpha}{(1-\beta_1)^2} \sum_{i=1}^{d}  \sum_{j=1}^{T}  \frac{ \sum_{l=1}^{T-j} \beta_{1}^{l} g_{j,i}^2 }{ \sqrt[p]{ j^{p/2-1}  v_{j,i}}}
  \\& \le  \frac{\alpha}{(1-\beta_1)^3} \sum_{i=1}^{d}  \sum_{j=1}^{T}  \frac{  g_{j,i}^2 }{ \sqrt[p]{ j^{p/2-1}  v_{j,i}}} .
  \end{align*}
  Applying Cauchy-Schwarz Inequality, we get
  \begin{align*}
   j^{p/2-1}  v_{j,i} & =  j^{p/2-1}  \cdot  \sum_{k=1}^{j} g_{k,i}^p 
   =  j^{p/2-2}  \cdot  j  \cdot  \sum_{k=1}^{j} ( g_{k,i}^{p/2}  )^2
  \\& \ge  j^{p/2-2}  \cdot  ( \sum_{k=1}^{j}  g_{k,i}^{p/2} )^2  =  ( j^{p/4-1}  \sum_{k=1}^{j}  g_{k,i}^{p/2} )^2
  \\&...
  \\& \ge ( \sum_{k=1}^{j}  g_{k,i}^{2} )^{p/2} .
  \end{align*}
  Then, we have
  \begin{align*}
  &\sum_{t=1}^{T}  \frac{\alpha_t}{1-\beta_1} \norm{m_t}_{V_t^{-1}}^2
  \\& \le  \frac{\alpha}{(1-\beta_1)^3} \sum_{i=1}^{d}  \sum_{j=1}^{T}  \frac{  g_{j,i}^2 }{ \sqrt{\sum_{k=1}^{j}  g_{k,i}^{2} }}
  \le \frac{2 \alpha}{(1-\beta_1)^3}   \sum_{i=1}^{d} \norm{g_{1:T,i} }_2 .
  \end{align*}
  In the last step, we apply Lemma 5 of~\citep{duchi2011adaptive} in inequality. let $ \beta_{1t} = \beta_1 \lambda^{t-1} $, we have
  \begin{align*}
  R(T) &\le 
  \frac{D_\infty^2}{2 (1-\beta_{1})} \sum_{i=1}^{d} \frac{b_T}{\alpha_T}  \sqrt[p]{v_{T,i}}
  + \frac{D_\infty^2}{2 } \sum_{t=1}^{T} \sum_{i=1}^{d}
  \frac{\beta_{1t}  V_{t-1,i} }{ (1-\beta_{1t}) \alpha_t }
  \\& + \frac{2 \alpha}{(1-\beta_1)^3}   \sum_{i=1}^{d} \norm{g_{1:T,i} }_2
  \\& \le \frac{D_\infty^2}{2 (1-\beta_{1})} T^{1/2 -1/p} \sum_{i=1}^{d} \norm{g_{1:T,i}}_p  
  + \frac{ \beta_1 D_\infty^2 G_\infty  }{2 (1-\beta_{1}) (1-\lambda)^2 } 
  \\& + \frac{2 \alpha}{(1-\beta_1)^3}   \sum_{i=1}^{d} \norm{g_{1:T,i} }_2 .
  \end{align*}
\end{proof}

\subsection{\textbf{Proof of Theorem~\ref{thm:main}} }
\begin{proof}
  Similar to the proof of~\ref{thm:equ-w}, we use the result of Theorem~\ref{thm:conv}, we have 
  \begin{align*}
  R(T) &\le 
  \frac{D_\infty^2}{2 (1-\beta_{1})} \sum_{i=1}^{d} \frac{b_T}{\alpha_T}  \sqrt[p]{v_{T,i}}
  + \frac{D_\infty^2}{2 } \sum_{t=1}^{T} \sum_{i=1}^{d}
  \frac{\beta_{1t}  V_{t-1,i} }{ (1-\beta_{1t}) \alpha_t }
  \\& + \sum_{t=1}^{T}  \frac{\alpha_t}{1-\beta_1} \norm{m_t}_{V_t^{-1}}^2 .
  \end{align*}
  First,we bound the last term of above inequality. Let $ \alpha_t = \frac{\alpha}{ \sqrt{t} } $, $ p_1=2 $,$ p_2=4 $, we have
 \begin{align*}
   &\sum_{t=1}^{T} {\alpha_t} \norm{m_t}_{V_t^{-1}}^2
   \\&  = \sum_{t=1}^{T-1} {\alpha_t} \norm{m_t}_{V_t^{-1}}^2 
   + \alpha \sqrt[4]{ \frac{ (1+T)T/2 }{T^2}}    \sum_{i=1}^{d} \frac{m_{T,i}^2}{ \sqrt[4]{v_{T,i}} } 
   \\& = \sum_{t=1}^{T-1} {\alpha_t} \norm{m_t}_{V_t^{-1}}^2 
   \\& + \alpha  \sqrt[4]{ \frac{(1+T)}{2T} } \sum_{i=1}^{d} \frac{
    (\sum_{j=1}^{T} \prod_{k=1}^{T-j} \beta_{1(T-k+1)} g_{j,i} )^2 
   }{
    \sqrt[4]{\sum_{j=1}^{T} j  \cdot  g_{j,i}^2}  
   }
   \\& \le \sum_{t=1}^{T-1} {\alpha_t} \norm{m_t}_{V_t^{-1}}^2 
   + \alpha \sum_{i=1}^{d} \frac{
    (\sum_{j=1}^{T}  \beta_{1}^{T-j})
    (\sum_{j=1}^{T}  \beta_{1}^{T-j} g_{j,i}^2 )
   }{
    \sqrt[4]{\sum_{j=1}^{T} j  \cdot  g_{j,i}^2}  
   }  
   \\& \le \sum_{t=1}^{T-1} {\alpha_t} \norm{m_t}_{V_t^{-1}}^2 
   + \frac{\alpha}{1-\beta_{1}} \sum_{i=1}^{d} \frac{
    \sum_{j=1}^{T}  \beta_{1}^{T-j} g_{j,i}^2
   }{
    \sqrt[4]{\sum_{j=1}^{T} j  \cdot  g_{j,i}^2}  
   }
  \\& \le \sum_{t=1}^{T-1} {\alpha_t} \norm{m_t}_{V_t^{-1}}^2 
  + \frac{\alpha}{1-\beta_{1}} \sum_{i=1}^{d} \sum_{j=1}^{T}  \frac{
    \beta_{1}^{T-j} g_{j,i}^2
  }{
    \sqrt[4]{\sum_{k=1}^{j} k  \cdot  g_{k,i}^2}  
  } .
\end{align*}
The first inequality follows Cauchy-Schwarz inequality. For the third inequality, we apply $ \sqrt[4]{\sum_{j=1}^{T} j  \cdot  g_{j,i}^2} \ge \sqrt[4]{\sum_{k=1}^{j} k  \cdot  g_{k,i}^2} $ when $ j \le T $. Then, we have
\begin{align*}
\sum_{t=1}^{T} {\alpha_t} \norm{m_t}_{V_t^{-1}}^2
\\&  \le \frac{\alpha}{1-\beta_{1}} \sum_{i=1}^{d} \sum_{j=1}^{T}  \frac{
  \sum_{l=1}^{T-j}  \beta_{1}^{l} g_{j,i}^2
}{
  \sqrt[4]{\sum_{k=1}^{j} k  \cdot  g_{k,i}^2}  
}
\\&  \le \frac{\alpha}{(1-\beta_{1})^2} \sum_{i=1}^{d} \sum_{j=1}^{T}  \frac{
  g_{j,i}^2
}{
  \sqrt[4]{\sum_{k=1}^{j} k  \cdot  g_{k,i}^2}  
} .
\end{align*}
Since $ g_i \le G_\infty $, applying Lemma~\ref{lem:sum-r}, we can get
\begin{equation}
\begin{aligned}
  \sum_{t=1}^{T} {\alpha_t} \norm{m_t}_{V_t^{-1}}^2
  &  \le \frac{\alpha d G_\infty}{(1-\beta_{1})^2} \sum_{i=1}^{d} \sqrt[4]{\sum_{j=1}^{T} j \cdot  g_{j,i}^2} .
\end{aligned}
\end{equation}
Finally,  let $ \beta_{1t} = \beta_1  \cdot  \lambda^{t-1}  $, since $ V_{t,i} \le \sqrt{G_\infty} $, we have,
\begin{equation}
\begin{aligned}
  &  R(T)  \le 
    \frac{D_\infty^2}{2 (1-\beta_{1})} \sum_{i=1}^{d} \sqrt[4]{\frac{(1+T)T/2}{T^2}}  \sqrt[4]{\sum_{j=1}^{T} j \cdot  g_{j,i}^2}
  \\&  + \frac{D_\infty^2}{2 } \sum_{t=1}^{T} \sum_{i=1}^{d}
  \frac{\beta_{1t}  V_{t-1,i} }{ (1-\beta_{1t}) \alpha_t } +
   + \frac{\alpha d G_\infty}{(1-\beta_{1})^2} \sum_{i=1}^{d} \sqrt[4]{\sum_{j=1}^{T} j \cdot  g_{j,i}^2}
  \\& \le 
   \frac{D_\infty^2}{2 (1-\beta_{1})} \sum_{i=1}^{d} \sqrt[4]{\sum_{j=1}^{T} j \cdot  g_{j,i}^2}
  \\&  + \frac{D_\infty^2}{2 } \sum_{t=1}^{T} \sum_{i=1}^{d}
  \frac{\beta_{1t}  V_{t-1,i} }{ (1-\beta_{1t}) \alpha_t }
   + \frac{\alpha d G_\infty}{(1-\beta_{1})^2} \sum_{i=1}^{d} \sqrt[4]{\sum_{j=1}^{T} j \cdot  g_{j,i}^2} .
\end{aligned}
\end{equation}
\end{proof}
\begin{minipage}{\columnwidth}
  \Biography{FCS-18457-author1}{
    Hui Zhong received the B.S. degree in Computer Science and Technology in 2016 from from the University of Science and Technology of China(USTC). He is currently a M.E. student in the School of Computer Science and Technology at USTC. His major research interests include machine learning and optimization.
  }
\end{minipage}
\begin{minipage}{\columnwidth}
  \Biography{FCS-18457-author2}{
  	Zaiyi Chen received the Ph.D. degree from University of Science and Technology of China (USTC), Hefei, China, in 2018. His major research interests include machine learning, optimization and sampling. He has published several papers in refereed conference proceedings, such as ICML'18, ICDM'16, SDM'15.
  }
\end{minipage}
\begin{minipage}{\columnwidth}
  \Biography{FCS-18457-author3}{
    Chuan Qin received the B.S degree in Computer Science and Technology from the University of Science and Technology of China (USTC) in 2015. He is currently working toward the PhD degree in the School of Computer Science and Technology, University of Science and Technology of China. His current research interests include natural language processing and recommender system.
  }
\end{minipage}
\begin{minipage}{\columnwidth}
  \Biography{FCS-18457-author4}{
    Zai Huang received the B.S. degree in Computer Science and Technology from University of Science and Technology of China (USTC) in 2016. He is currently pursuing the M.S. degree in Computer Application Technology from USTC. His current research interests include data mining and machine learning.
  }
\end{minipage}
\begin{minipage}{\columnwidth}
  \Biography{FCS-18457-author5}{
    Vincent W. Zheng is an Adjunct Senior Research Scientist at Advanced Digital Sciences Center (ADSC), Singapore. He received his Ph.D. degree from the Hong Kong University of Science and Technology in 2011. His research interests focus on mining with heterogeneous and structured data. He is the Associate Editor of Cognitive Computation. He has served as PCs in many leading data mining and artificial intelligence conferences such as KDD, IJCAI, AAAI, WWW, WSDM. He has published over 60 papers in the refereed conferences, journals and book chapters. He is a member of AAAI and ACM.}
\\\\\\
\end{minipage}
\begin{minipage}{\columnwidth}
  \Biography{FCS-18457-author6}{
    Tong Xu received the Ph.D. degree in University of Science and Technology of China (USTC), Hefei, China, in 2016. He is currently working as an Associate Researcher of the Anhui Province Key Laboratory of Big Data Analysis and Application, USTC. He has authored 20+ journal and conference papers in the fields of social network and social media analysis, including KDD, AAAI, ICDM, SDM, etc.}
\end{minipage}
\begin{minipage}{\columnwidth}
  \Biography{FCS-18457-author7}{
    Enhong Chen is a professor and vice dean of the School of Computer Science at USTC. He received the Ph.D. degree from USTC. His general area of research includes data mining and machine learning, social network analysis and recommender systems. He has published more than 100 papers in refereed conferences and journals, including IEEE Trans. KDE, IEEE Trans. MC, KDD, ICDM, NIPS, and CIKM. He was on program committees of numerous conferences including KDD, ICDM, SDM. His research is supported by the National Science Foundation for Distinguished Young Scholars of China. He is a senior member of the IEEE.
  }
\end{minipage}
 
\end{document}